\documentclass[conference]{IEEEtran}
\IEEEoverridecommandlockouts

\usepackage{cite}
\usepackage{amsmath,amssymb,amsfonts}
\usepackage{graphicx}
\usepackage{textcomp}
\usepackage{xcolor}
\usepackage{booktabs}
\usepackage{multirow}
\usepackage{array}
\usepackage{tabularx}
\usepackage{url}
\usepackage{microtype}
\usepackage{nicefrac}
\usepackage{xspace}
\usepackage[hidelinks]{hyperref}
\usepackage{tikz}
\usetikzlibrary{arrows.meta, calc, positioning}
\usepackage{placeins}

\graphicspath{{figures/}}

\newcommand{\PCSP}{\textsc{pcsp}\xspace}
\newcommand{\PCSPD}{\textsc{pcsp-d}\xspace}
\newcommand{\eg}{\textit{e.g.}\xspace}

\title{One Policy, Infinite NPCs:\\
Persona-Traceable Shared RL Policies\\
for Scalable Game Agents}

\author{%
  \IEEEauthorblockN{Yoosung Hong\thanks{Project page: \url{https://github.com/yoosunghong/pcsp}.}}
  \IEEEauthorblockA{Independent Researcher\\
    \url{yoosunghong.main@gmail.com}}
}

\begin{document}
\maketitle

\begin{abstract}
On a 300-persona life-simulation benchmark, \PCSP achieves compositional
zero-shot persona identification (unseen-occupation held-out) up to
17$\times$ above chance, Spearman $\rho\!\approx\!0.73$ semantic-behavioral
alignment, and 22$\times$ faster inference than an LLM-as-policy baseline. Life simulation games require hundreds to thousands of
non-player characters (NPCs) that behave consistently with distinct personalities
while remaining controllable through designer-authored natural language.
Hand-authored behavior trees, per-character RL policies, unsupervised skill
discovery, and per-step LLM controllers each fail on one or more deployment
constraints: persona consistency, natural-language controllability, zero-shot
generalization, and real-time inference.
We introduce \PCSP (Persona-Conditioned Shared Policy), a single reinforcement
learning policy conditioned on frozen LLM embeddings of free-form persona
descriptions. \PCSP combines once-per-NPC persona encoding, low-rank persona
projection, neural persona conditioning, and a PPO + InfoNCE consistency + KL
diversity training objective. Across three \PCSPD{} (formerly \textit{Mini-Inzoi}) experimental settings,
including a richer 20-action v3 ontology, ablations show that the InfoNCE
trajectory-consistency objective is load-bearing: removing it collapses
zero-shot persona identification to chance even when task reward is preserved or
improved. External validation on three \textit{Melting Pot
2.4.0}~\cite{leibo2021meltingpot} substrates spanning commons-pool, public-good,
and dyadic-matrix social dilemmas (\texttt{commons\_harvest\_\_open},
\texttt{clean\_up}, \texttt{prisoners\_dilemma\_in\_the\_matrix\_\_repeated})
confirms that the \S\ref{sec:method} method produces persona-conditioned
behavioral divergence in multi-agent strategic substrates and that the
consistency-loss ablation collapses trajectory$\to$persona retrieval to chance
in every substrate while leaving (or inflating) pairwise action-KL.
We distinguish two senses of held-out evaluation: \emph{compositional zero-shot}
(unseen occupation $\times$ archetype crosses within the trained
persona-space coverage; the regime of Layer 1 and Layer 3), and
\emph{vocabulary-expansion held-out} (new persona tokens whose embeddings
lie inside the convex hull of training embeddings but were never present at
training; the regime where Layer 2 still fails, top-1 $=0$, and which we
report as an open problem). A UE5
deployment reproduces the in-engine persona-conditioning ablation at 64 agents
with a 1.7\% failure rate and held-out zero-shot generalization at 0.04\%
failure, showing that the sub-frame inference profile and ablation structure
survive the move into a commercial game engine. These results provide empirical
evidence that shared RL policies can support scalable, real-time,
persona-conditioned NPC control and that trajectory traceability is central to
the method.
\end{abstract}


\section{Introduction}
\label{sec:intro}

Modern life simulation games---\textit{The Sims}, \textit{Animal Crossing},
\textit{inZOI}, and emerging open-world titles---place NPCs at the center of
the player experience. For these games to feel alive, each NPC must behave
consistently with a distinct personality: the gregarious chef and the reclusive
artist should pursue the same biological needs through recognizably different
activity patterns, social approaches, and daily rhythms. At scale---hundreds to
thousands of NPCs per game world---this creates a fundamental challenge that
current game AI architectures were not designed to address.

\subsection{The NPC Personalization Scaling Gap}

\paragraph{Behavior trees} remain the industrial standard~\cite{yannakakis2018ai,colledanchise2018bt}.
A skilled designer hand-crafts decision logic for each character archetype.
This produces believable, predictable NPCs, but authoring cost scales linearly
with character count, and trees are brittle outside authored scenarios.
Ten thousand distinct NPC personalities require ten thousand hand-crafted trees.

\paragraph{Per-NPC RL} appears to offer automation: train a separate policy for
each persona. In principle this allows unlimited characters with arbitrary
behaviors. In practice, memory and training cost also scale linearly, and each
new persona requires a full retraining cycle---a prohibitive expense for
live-service games that regularly introduce new characters.

\paragraph{LLM-as-policy} methods~\cite{park2023generative,wang2023voyager,yao2023react,shinn2023reflexion,wei2022chain}
achieve rich, natural-language-grounded persona expression by querying a language
model at every decision step. This works well in offline or turn-based contexts,
but even our local Qwen3-1.7B LLM-as-policy baseline requires 43.7\,ms per
decision step (Table~\ref{tab:results_v1}),
which exceeds the 16--33\,ms per-frame budget of real-time game simulations.
Generative Agents~\cite{park2023generative} sidestep this by operating on
minute-resolution plans rather than per-frame actions, which suits narrative
simulation but cannot drive moment-to-moment NPC movement and activity selection.

\paragraph{Unsupervised skill discovery} (DIAYN~\cite{eysenbach2018diayn},
CIC~\cite{laskin2022cic}), often paired with intrinsic-motivation
objectives~\cite{pathak2017curiosity}, learns a shared policy conditioned on latent skill codes,
achieving behavioral diversity without per-character training. But the codes carry
no semantic content. A game designer cannot specify ``this NPC should be agreeable
and fitness-oriented'' by selecting a latent index.
Natural-language controllability is not part of the design.

\subsection{Persona-Conditioned Shared Policies}

We address this scaling problem with a different decomposition: \emph{compute a
rich persona representation once per NPC, then condition a single lightweight
policy on that representation at every step}.

The key enabler is the modern LLM embedding space. A frozen language model maps
free-form persona descriptions to dense vectors that capture semantic
relationships among personalities. A shared policy conditioned on these
embeddings can generalize to \emph{any} persona a designer writes, without
retraining, because the continuous embedding space provides coverage of the
entire personality manifold. At inference time, the LLM is called once per NPC
lifetime; the policy network---at most a few hundred kilobytes---runs at full
game speed.

This approach, which we call \emph{persona-conditioned shared
policies}, remains underexplored in game AI. This paper presents a concrete
implementation, \PCSP, and evaluates when persona-conditioned behavior remains
recoverable from generated trajectories. The central empirical question is not
whether a policy can receive persona text as input, but whether its trajectories
preserve enough persona signal to support zero-shot identification, semantic
alignment, and real-time deployment.

\subsection{Paper Contributions}

\begin{enumerate}
  \setlength\itemsep{2pt}
  \item \textbf{Method.} \PCSP{}---a shared policy conditioned on frozen LLM
    persona embeddings via a low-rank persona projection (LoRA-style matrix
    factorization used as a standalone projection layer, not as a parallel
    adapter on a pretrained weight) and FiLM/concat fusion,
    co-trained with PPO, an InfoNCE trajectory-consistency objective, and
    KL diversity regularization (\S\ref{sec:method}).
  \item \textbf{Three-layer validation methodology.} We argue that
    persona-conditioned agents require \emph{separated} validation of
    mechanism, generalization, and deployment, and instantiate this with a
    controlled diagnostic substrate (\PCSPD), an external multi-agent RL
    substrate (Melting Pot), and a realtime UE5 engine deployment
    (\S\ref{sec:layers}).
  \item \textbf{Mechanistic finding (Layer 1).} Under controlled conditions,
    the InfoNCE consistency term is causally responsible for trajectory-level
    persona recoverability: removing it preserves task reward but collapses
    zero-shot persona identification to chance across three independent
    environment instantiations of \PCSPD (\S\ref{sec:evidence}).
  \item \textbf{External generalization (Layer 2).} The same method, without
    algorithmic modification beyond a CNN observation front-end, transfers to
    three Melting Pot social-dilemma substrates spanning commons-pool,
    public-good, and dyadic-matrix structures
    (\texttt{commons\_harvest\_\_open}, \texttt{clean\_up},
    \texttt{prisoners\_dilemma\_\dots\_repeated}); the consistency-loss
    ablation collapses trajectory$\to$persona retrieval to chance in all
    three while leaving (or inflating) pairwise action-KL
    (\S\ref{sec:meltingpot}). Retrieval over the full $12$-persona vocabulary
    remains $3.4$--$4.9\times$ chance top-1, and a CU $\to$ CH cross-substrate
    transfer of the persona projection and trajectory encoder retrieves at
    $1.79\times$ chance top-1 (\S\ref{sec:mp_transfer},
    Tab.~\ref{tab:mp_transfer}).
  \item \textbf{Deployment finding (Layer 3).} A frozen Layer-1 checkpoint
    deployed in UE5 sustains 64 concurrent persona-conditioned agents at
    realtime with 1.7\% failure, generalizes to 60 held-out personas at
    0.04\% failure, and \emph{reproduces the InfoNCE ablation in-engine}
    (matched-persona symmetric KL of $1.79$\,nats between full and
    \texttt{no\_consist} checkpoints, reward $1{,}423.5$ vs.\ $1{,}079.8$
    on the same map and persona set), establishing that the consistency
    objective is load-bearing under engine-side contention
    (\S\ref{sec:ue5}).
  \item \textbf{Reproducibility.} Open ONNX checkpoints, three-layer
    benchmark code, UE5 plugin, and trajectory-annotation harness.
\end{enumerate}

\section{Why Current Paradigms Fall Short}
\label{sec:paradigms}

Table~\ref{tab:paradigms} evaluates six paradigms against the four axes that
matter for practical life-simulation NPC deployment.

\begin{table}[t]
  \centering
  \small
  \caption{Paradigm comparison. \checkmark\,= satisfied, $\times$ = not
    satisfied, $\triangle$ = partially satisfied. The \PCSP{} row reports
    satisfaction \emph{in our evaluated settings} (the three-layer stack of
    \S\ref{sec:layers}); open issues such as Melting Pot vocabulary-expansion
    held-out recovery (\S\ref{sec:mp_transfer}), broader human believability,
    and open-world generalization beyond Layer 3 are discussed in
    \S\ref{sec:limits}.}
  \label{tab:paradigms}
  \setlength{\tabcolsep}{3pt}
  \begin{tabular}{lcccc}
    \toprule
    Paradigm &
      \parbox{1.4cm}{\centering Persona\\consist.} &
      \parbox{1.2cm}{\centering NL\\control.} &
      \parbox{1.3cm}{\centering Zero-shot\\gen.} &
      \parbox{1.1cm}{\centering Real-time\\($<$5\,ms)} \\
    \midrule
    Behavior trees           & $\triangle$ & $\times$   & $\times$   & \checkmark \\
    Goal-cond.\ RL           & $\times$    & $\times$   & $\times$   & \checkmark \\
    Lang-cond.\ RL           & $\triangle$ & \checkmark & $\triangle$& \checkmark \\
    Skill discovery          & $\triangle$ & $\times$   & $\times$   & \checkmark \\
    Per-NPC RL               & \checkmark  & $\times$   & $\times$   & \checkmark \\
    LLM-as-policy            & \checkmark  & \checkmark & \checkmark & $\times$   \\
    \midrule
    \textbf{\PCSP (ours)}    & \checkmark  & \checkmark & \checkmark & \checkmark \\
    \bottomrule
  \end{tabular}
\end{table}

\paragraph{Goal-conditioned RL.}
UVFA~\cite{schaul2015uvfa} and HER~\cite{andrychowicz2017her} condition policies
on goal \emph{states}---what to achieve. Persona describes \emph{how to behave}:
two NPCs with identical needs but different personalities should fulfill them
through different activity sequences. Goal conditioning addresses the wrong axis
of variation and provides no natural-language interface.

\paragraph{Language-conditioned RL.}
BabyAI~\cite{chevalier2019babyai} conditions on natural-language
\emph{instructions} that change each episode.
Persona is a stable trait persisting over the NPC's lifetime; treating it as an
episode-level instruction discards the long-horizon consistency constraint.
Decision Transformer~\cite{chen2021decision} and similar sequence-modeling
approaches condition behavior on past returns or goals, but not on static identity
descriptors that must generalize zero-shot.

\paragraph{Unsupervised skill discovery.}
DIAYN~\cite{eysenbach2018diayn} and CIC~\cite{laskin2022cic} achieve behavioral
diversity via learned latent codes, but the codes carry no semantic content.
Generalizing to a new designer-specified persona requires solving an inverse
problem: find the latent code for ``an introverted accountant who prioritizes
learning.'' This is not supported by design.

\paragraph{LLM-as-policy.}
Generative Agents~\cite{park2023generative}, Voyager~\cite{wang2023voyager},
and ReAct~\cite{yao2023react} produce rich, persona-consistent behavior but
remain expensive when used as per-step controllers. In our benchmark, the
Qwen3-1.7B LLM-as-policy baseline requires 43.7\,ms per decision step
(Table~\ref{tab:results_v1}), already above a 16--33\,ms frame budget.
Techniques such as action caching, plan reuse, and smaller distilled models can
reduce latency, but the fundamental tension---rich language reasoning versus
real-time step budgets---remains. Generalist agents such as
Gato~\cite{reed2022gato} demonstrate that a single model can handle diverse
tasks, but inference cost is similar. Large-scale self-play agents such as
AlphaStar~\cite{vinyals2019alphastar}, OpenAI Five~\cite{berner2019dota},
and AlphaZero~\cite{silver2018alphazero} establish that RL can master complex
games, but they specialize a single policy to a single game rather than
expressing a vocabulary of personas within one world.

\paragraph{Summary.}
No existing paradigm satisfies all four axes. The design space between
``fast but no NL control'' and ``NL control but too slow'' has not been
systematically explored. Persona-conditioned shared policies are a concrete
proposal for how to close this gap.

\section{\PCSP{}: Persona-Conditioned Shared Policy}
\label{sec:method}

We present \PCSP as one specific point in the design space of persona-conditioned
shared policies. The components below are motivated by ablation evidence
but are not canonical; they define the experimental system evaluated across the
three validation layers in \S\ref{sec:layers}. The method itself is
\emph{environment-agnostic}: a frozen embedding plus a low-rank persona projection,
a shared policy with a conditioning fusion, and a PPO+InfoNCE+KL co-training
objective. Environment, observation, action, and reward specifications are
deferred to \S\ref{sec:layers} together with the layer they instantiate.

\subsection{Persona Encoding}

Given persona text $p$, we compute a frozen LLM embedding
$\hat{e}_p \in \mathbb{R}^{1024}$ using the Transformer-based~\cite{vaswani2017attention}
Qwen3-0.6B-Embedding~\cite{qwen3embed}
(last-token pooling, L2-normalized); contrastively-trained sentence encoders
such as Sentence-BERT~\cite{reimers2019sbert} provide our B3 baseline below.
This is computed once per NPC lifetime (14.6\,ms/persona), imposing negligible
runtime cost.

A learned low-rank persona projection ($r\!=\!16$,
$\text{lr}\!=\!10^{-4}$) maps $\hat{e}_p \to e_p \in \mathbb{R}^{64}$
through a low-dimensional projection layer whose matrix-factorization
structure is inspired by LoRA~\cite{hu2022lora}---we do not add a parallel
$BA$ adapter to a pretrained weight matrix in the Qwen3 backbone; rather, we
factor the standalone $1024\!\to\!64$ persona projection as a rank-$16$
$BA$ to keep the learnable persona pathway small while leaving the embedding
model frozen:
\[
\begin{aligned}
  e_p &= \operatorname{norm}\!\left(\alpha\, B A \hat{e}_p\right), \\
  A &\in \mathbb{R}^{16\times1024},\quad
  B \in \mathbb{R}^{64\times16},\quad
  \alpha = 64^{-1/2}.
\end{aligned}
\]
The projection is necessary: raw Qwen3 embeddings over-represent occupational
similarity relative to personality similarity (salesperson$\leftrightarrow$executive:
0.61 vs.\ trainer$\leftrightarrow$blogger: 0.33).
After low-rank projection training, Spearman $\rho$ between projected persona
distance and behavioral KL increases from 0.384 to 0.728
(\S\ref{sec:evidence}).

\subsection{Persona-Conditioned Shared Policy}

The shared policy $\pi_\theta(a \mid s, e_p)$ is a 3-hidden-layer MLP
(256-256-128) with an output head sized to the environment action space.
Our reference
implementation uses FiLM~\cite{perez2018film} to inject the persona signal at
every hidden layer:
\[
  h_\ell' = \gamma_\ell(e_p) \odot h_\ell + \beta_\ell(e_p),
\]
with small linear networks
$\gamma_\ell, \beta_\ell : \mathbb{R}^{64} \to \mathbb{R}^{d_\ell}$,
where $d_\ell$ is the hidden width of layer $\ell$.
An analogous FiLM-conditioned value network $V_\psi(s, e_p)$ serves as the
critic. Total trainable parameters: 207K (excluding the frozen LLM).
We treat the choice of conditioning mechanism as an implementation detail
rather than a contribution: the v3 zero-shot ablations in
\S\ref{sec:evidence} show that input-concatenation conditioning is competitive
with FiLM and in fact generalizes more strongly to unseen occupations. The
load-bearing component is the consistency objective introduced next, which is
what makes persona signal recoverable from trajectories regardless of how
the policy is conditioned.

\subsection{Co-Training Objective}

\begin{equation}
  \mathcal{L} = \mathcal{L}_{\text{PPO}} + \lambda_1 \mathcal{L}_{\text{consist}}
    + \lambda_2 \mathcal{L}_{\text{diverse}},
\end{equation}
with $\lambda_1\!=\!0.5$, $\lambda_2\!=\!0.1$ (PPO: $\gamma\!=\!0.99$,
$\epsilon\!=\!0.2$, GAE-$\lambda\!=\!0.95$, batch\,$=\!2048$~\cite{schulman2017ppo};
cooperative multi-agent PPO variants~\cite{yu2022mappo,dewitt2020ippo} share the
same per-agent update used here).

\textbf{Consistency loss.}
A 2-layer GRU trajectory encoder $g_\eta(\tau) \in \mathbb{R}^{64}$
maps episode trajectories to L2-normalized embeddings.
InfoNCE~\cite{oord2018cpc} ($T\!=\!0.07$) aligns trajectory embeddings with
their originating persona embeddings:
\begin{equation}
  \mathcal{L}_{\text{consist}} = -\log
  \frac{\exp\!\bigl(\text{sim}(g_\eta(\tau), e_p)/T\bigr)}
       {\sum_{p' \in \mathcal{B}} \exp\!\bigl(\text{sim}(g_\eta(\tau), e_{p'})/T\bigr)}.
\end{equation}
This prevents the policy from producing trajectories that cannot be traced back
to their conditioning persona.

\textbf{Diversity loss.}
We maximize the expected KL divergence between policy distributions induced by
different personas at shared states:
\begin{equation}
  \mathcal{L}_{\text{diverse}} =
    -\mathbb{E}_{s,\, p \neq p'}\!\bigl[
      D_{\mathrm{KL}}\!\bigl(\pi_\theta(\cdot \mid s, e_p)
        \;\|\; \pi_\theta(\cdot \mid s, e_{p'})\bigr)
    \bigr],
\end{equation}
computed as a batched forward pass over $n\!=\!8$ sampled personas at 32 states.
This discourages behavioral collapse toward a persona-agnostic average.

\begin{figure*}[t]
  \centering
  \includegraphics[width=0.98\textwidth]{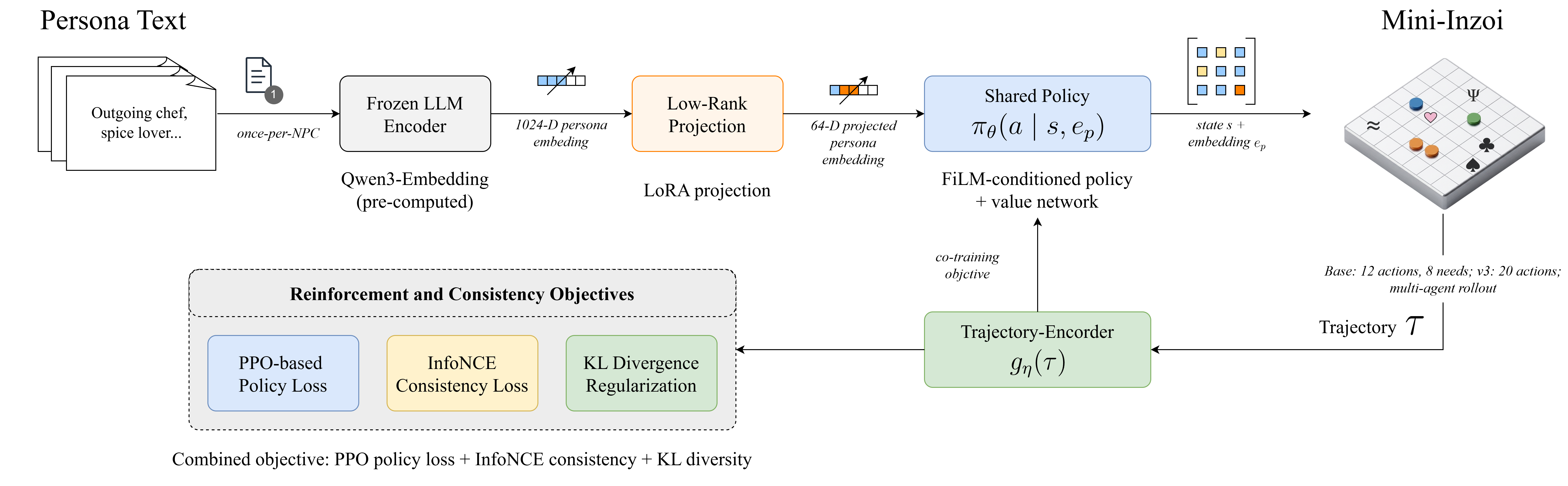}
  \caption{%
    \textbf{\PCSP pipeline and training objectives.}
    Persona text is encoded once per NPC with a frozen Qwen3 embedding model,
    adapted through a low-rank projection, and consumed by a shared
    persona-conditioned policy during \PCSPD{} rollout.
    The trajectory encoder provides the InfoNCE consistency signal, while the
    policy is optimized with PPO and KL diversity regularization.
    The rightmost \PCSPD{} label depicts the base 12-action/8-need
    instantiation; the v3 experiments expand the action ontology to 20 actions.}
  \label{fig:system}
\end{figure*}

\section{Evaluation Strategy: A Three-Layer Validation Stack}
\label{sec:layers}

A persona-conditioned policy must answer three different questions:
(i) is persona signal causally responsible for trajectory variation;
(ii) does the same method survive a different environment, observation
geometry, and reward structure; and (iii) does the policy hold up under the
asynchrony, contention, and runtime constraints of a commercial game engine.
A single environment cannot answer all three: the conditions that make
causal isolation possible (full observability, small action space, short
episodes) are mutually exclusive with the conditions that test deployment
realism. We therefore validate \PCSP on three deliberately different layers
and report each layer against the question it can actually answer
(Table~\ref{tab:layer_matrix}).

\begin{figure}[t]
  \centering
  \includegraphics[width=\linewidth]{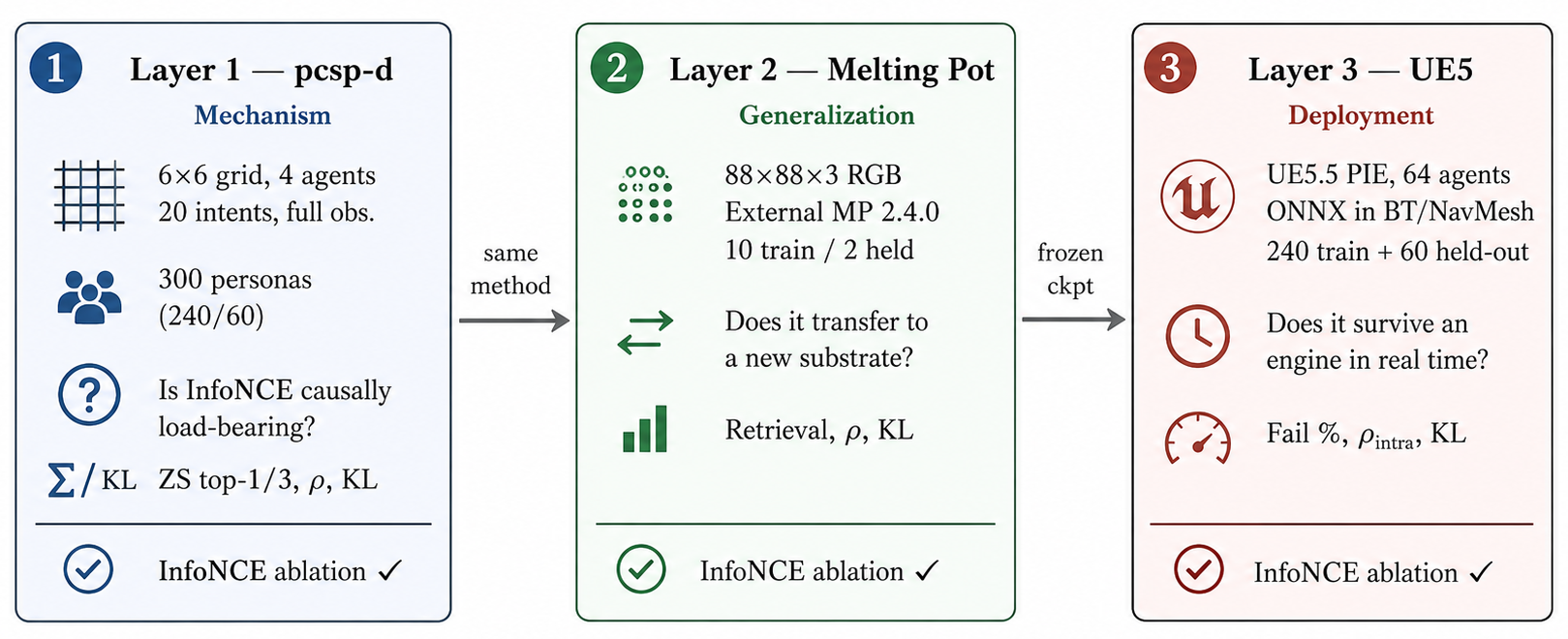}
  \caption{%
    \textbf{Three-layer validation stack.} Each layer is selected for the
    question it can isolate; the InfoNCE consistency term is ablated in all
    three. Together they cover mechanism, generalization, and deployment.}
  \label{fig:three_layer}
\end{figure}

\begin{table*}[t]
  \centering
  \small
  \caption{%
    \textbf{Three-layer validation matrix.} Per-layer evaluation surface,
    persona partition, and whether the InfoNCE consistency term is ablated
    inside that layer.}
  \label{tab:layer_matrix}
  \setlength{\tabcolsep}{4pt}
  \renewcommand{\arraystretch}{1.08}
  \begin{tabularx}{\textwidth}{c
      >{\raggedright\arraybackslash}X
      >{\raggedright\arraybackslash}X
      >{\raggedright\arraybackslash}p{0.17\textwidth}
      >{\raggedright\arraybackslash}X
      c}
    \toprule
    Layer & Substrate & Question answered & Personas (train/eval) & Key metrics & InfoNCE abl.\ \\
    \midrule
    1 & \PCSPD{} 6$\times$6 grid, 4 agents (\S\ref{sec:layer1env})
      & Is InfoNCE causally load-bearing?
      & 240 / 60 held-out
      & ZS top-1/3, $\rho$, pairwise KL
      & \checkmark \\
    2 & Melting Pot 2.4.0, RGB 88$\times$88, 7--8 agents (\S\ref{sec:meltingpot})
      & Does the same method transfer to a new substrate?
      & 10 / 2 held-out%
        \footnote{Persona budget mirrors the \PCSPD{}-v2 controlled split; the 240/60 protocol is the planned T1.2 expansion.}
      & Trajectory retrieval top-1/3, $\rho$, pairwise KL
      & \checkmark \\
    3 & UE5.5 hybrid stack, 64 agents (\S\ref{sec:ue5})
      & Does the consistency finding survive an engine?
      & 240 / 60 held-out
      & Fail \%, throughput, $\rho_{\text{intra}}$, sym.\ KL
      & \checkmark (\S\ref{sec:ue5}) \\
    \bottomrule
  \end{tabularx}
\end{table*}

\subsection{Layer 1 — \PCSPD{}: a controlled substrate for persona-traceability analysis}
\label{sec:layer1env}

Validating that a policy's behavior is causally traceable to its conditioning
embedding requires an environment where (i) every state transition is fully
observable, (ii) the action space is small enough to compute exact trajectory
distributions and KL divergences, (iii) the reward composition is controllable
so that the contribution of any persona-aligned shaping term can be isolated
or removed, and (iv) episode length is short enough to run thousands of held-out
personas. No commercial or photorealistic environment satisfies all four
simultaneously. We therefore construct \PCSPD{} (\PCSP-Diagnostic; formerly
\textit{Mini-Inzoi}), an intentionally minimal PettingZoo~\cite{terry2021pettingzoo} AEC substrate
(6$\times$6 grid, 4 agents, 20 discrete intents over 8 bio-social needs) whose
role in this paper is \emph{not} to demonstrate behavioral realism---that is
the role of Layers 2 and 3---but to expose the InfoNCE consistency term to
controlled ablation under conditions where every causal pathway from persona
to trajectory is analytically observable. We treat \PCSPD{} as a microscope,
not a world.

\textbf{Environment.}
\PCSPD{} has 8 bio-social needs (hunger, sleep, social, leisure, hygiene,
fitness, work, learning) and 12 discrete actions in the v1/v2 base ontology
(8 activity + 4 movement); the primary v3 instantiation expands this to 20
discrete intents (16 activity + 4 movement) for finer behavioral granularity.
Observation $s\!\in\!\mathbb{R}^{20}$ (v1/v2) or $\mathbb{R}^{33}$ (v3):
position, time-of-day, needs, summaries of the other agents, and (v3) location
affordance, social context, and routine regularity.
Reward combines needs satisfaction, persona-aligned activity bonuses ($+0.5$
for preferred actions), and social bonuses scaled by Big Five
compatibility~\cite{mccrae1992introduction,goldberg1990bigfive}
($0.2 + 0.3 \times \text{cos-sim}(\text{BF}_i, \text{BF}_j)$).
The v1/v2 base ontology uses only needs-driven and persona-agnostic social
shaping; the v3 instantiation additionally adds a per-action persona-style
term (\S\ref{sec:evidence}) that is \emph{declared as part of the
\PCSPD-v3 environment}, not as a method component, and that the InfoNCE
ablation (\S\ref{sec:evidence}) is run against without modification so that
any persona-recoverability collapse cannot be attributed to removing this
shaping. The mechanistic claim of the paper---that the InfoNCE consistency
loss is what makes trajectories traceable to their conditioning persona---is
therefore demonstrated under \emph{both} the persona-agnostic-reward regime
(v1/v2) and the persona-style-reward regime (v3): the no\_consist ablation
collapses zero-shot identification to chance in every case.

\textbf{Persona dataset.}
300 persona texts are generated from 15 Big Five archetypes $\times$ 20
occupations (240 train / 60 zero-shot test).
Each text is a 2--3 sentence natural-language description that an NPC designer
might write for a new character.

\subsection{Layer 2 — Cross-substrate generalization on Melting Pot}
\label{sec:layer2env}

Layer 1 cannot test whether the method survives a different observation
geometry, action ontology, or social-dilemma structure. Layer 2 applies
\PCSP{} unchanged to Melting Pot 2.4.0~\cite{leibo2021meltingpot}: an
external multi-agent RL substrate with 88$\times$88$\times$3 RGB
observations and an 8-action ontology. The same protocol is applied
without algorithmic modification across three substrates spanning distinct
social-dilemma structures---\texttt{commons\_harvest\_\_open} (commons-pool),
\texttt{clean\_up} (public-good), and
\texttt{prisoners\_dilemma\_\_in\_the\_matrix\_\_repeated}
(dyadic matrix-game)---reported in \S\ref{sec:meltingpot} (Tab.~\ref{tab:mp_multi}).

\subsection{Layer 3 — Realtime deployment in Unreal Engine 5}
\label{sec:layer3env}

A persona-conditioned policy is only meaningful if it survives the
engineering pressure of a real game engine: asynchronous tick rates, NavMesh
contention, BT failure recovery, ONNX runtime constraints, and shared world
state. Layer 3 deploys a \emph{frozen} Layer-1 checkpoint into UE5.5 via a
hybrid intent stack and asks three questions that Layers 1 and 2 cannot
answer: (i) does the policy meet a realtime wall-budget at deployment scale;
(ii) does the InfoNCE finding survive engine-side contention; and (iii) do
personas maintain identity over horizons far longer than the training episode.

\subsection{Per-layer evaluation protocols and metrics}
\label{sec:evalprotocol}

We evaluate persona-conditioned NPC control with metrics that separate task
performance from persona fidelity.

\paragraph{Persona identification accuracy.}
Given a trajectory, an automated $k$-NN classifier identifies which held-out
persona generated it. We report zero-shot accuracy on held-out personas, with
random chance determined by the number of test personas.

\paragraph{Behavioral diversity.}
Mean pairwise KL divergence between action distributions conditioned on different
persona embeddings measures whether policies collapse to persona-agnostic
behavior despite high task reward.

\paragraph{Semantic-behavioral alignment.}
Spearman $\rho$ between projected persona distances and pairwise behavioral KL
divergences measures whether semantic distances in persona space are reflected
in behavior space. The projected embeddings are L2-normalized, so Euclidean
distance is monotonic with cosine distance. High $\rho$ with low KL variance can
indicate alignment without meaningful diversity, so both quantities are reported
where available.

\paragraph{Inference latency.}
We report GPU ms/step at batch size 1. Real-time game integration typically
requires per-step control below a 16--33\,ms frame budget.

\paragraph{Rich trajectory observability.}
Identification and believability scores depend on what a trace exposes.
Coarse action sequences can compress out temporal, spatial, social, and
stylistic dimensions that personality traits modulate. We therefore render
rollouts with time-of-day, location, nearby agents, action style, and short-form
event semantics for human-facing evaluation artifacts, and treat coarse-versus-
rich trace comparison as part of the evaluation protocol.

\section{Layer 1 Results: Mechanistic Validation}
\label{sec:evidence}

We evaluate \PCSP on four \PCSPD{} instantiations. v3 is the primary
experiment because it exposes a richer action ontology for
persona-conditioned behavior; v1, v2, and v3-large (all deferred to
App.~\ref{app:v1v2}) replicate the central InfoNCE finding at the original
12-action base scale, at a 12$\times$12/16-agent 12-action expansion, and
at a 12$\times$12/16-agent \emph{20-action} expansion at 500 personas,
establishing that the load-bearing component is invariant to grid size,
agent count, persona-set size, and action ontology.

\subsection{Experimental Setup}

\textbf{v3 (richer action ontology).}
\PCSPD-v3 keeps the 6$\times$6/4-agent base scale but expands the action
space to \textbf{20 flat-discrete actions} (16 activity + 4 movement) for
finer behavioral granularity (e.g., \texttt{focused\_work} vs.\
\texttt{planning\_work}, \texttt{rest\_alone} vs.\ \texttt{rest\_with\_others},
\texttt{eat\_quick} vs.\ \texttt{eat\_slow}).
Observations expand to 33 dimensions, adding location-affordance one-hot
(8), social-context features (3), and a routine-regularity signal (2).
Reward shaping adds a per-action persona-style term
$r_{\text{style}} = 0.3 \cdot \cos(\mathbf{p}_{\text{BF}}, \mathbf{s}_a)$
that aligns each action with a hand-authored Big-Five style profile.
Same 240/60 unseen-occupation split as v1; 300 PPO iterations.

\textbf{v1 (base scale).}
\PCSPD-v1: 6$\times$6, 4 agents, 300 personas (240 train / 60 zero-shot test).
The held-out 60 personas comprise \emph{four entire occupations} (HR manager,
professor, data scientist, chef) crossed with all 15 Big Five archetypes, so
this evaluates \textbf{unseen-occupation compositional generalization}: the
target occupation never appears in training, only the personality structure
does.
All models train for 300 PPO iterations ($\approx$1.9M steps) on an NVIDIA
RTX 6000 Ada (49\,GB VRAM).
Baselines: \textbf{B1} No-Persona PPO; \textbf{B3} SBERT-conditioned policy
(\texttt{all-MiniLM-L6-v2}, 384-dim, frozen); \textbf{B4} DIAYN (random 64-dim);
\textbf{B5} LLM-as-policy (Qwen3-1.7B).

\textbf{v2 (expanded scale).}
\PCSPD-v2: \textbf{12$\times$12, 16 agents, 500 personas} (400 train / 100
zero-shot test), 200 PPO iterations.
Same hyperparameters and evaluation protocol as v1.
The observation dimension expands to 56 (position 2, time 1, needs 8,
15 other agents $\times$ 3).
Random zero-shot chance is 1.0\% (1/100 test personas).

\textbf{v3-large (richer ontology at the expanded scale).}
\PCSPD-v3-large combines the v3 20-action ontology, affordance/social/routine
observation, and style reward with the v2 footprint:
\textbf{12$\times$12, 16 agents, 500 personas} (400 train / 100 zero-shot
test), obs.\ dim.\ 69, 300 PPO iterations, 3 seeds per arm.
The role of this instantiation is to test whether the v3 InfoNCE finding
holds when grid, agent count, and persona-set size all expand simultaneously
under the richer action ontology. Random zero-shot chance is 1.0\%.

\subsection{Results}

\begin{figure*}[!t]
  \centering
  \includegraphics[width=0.9\textwidth]{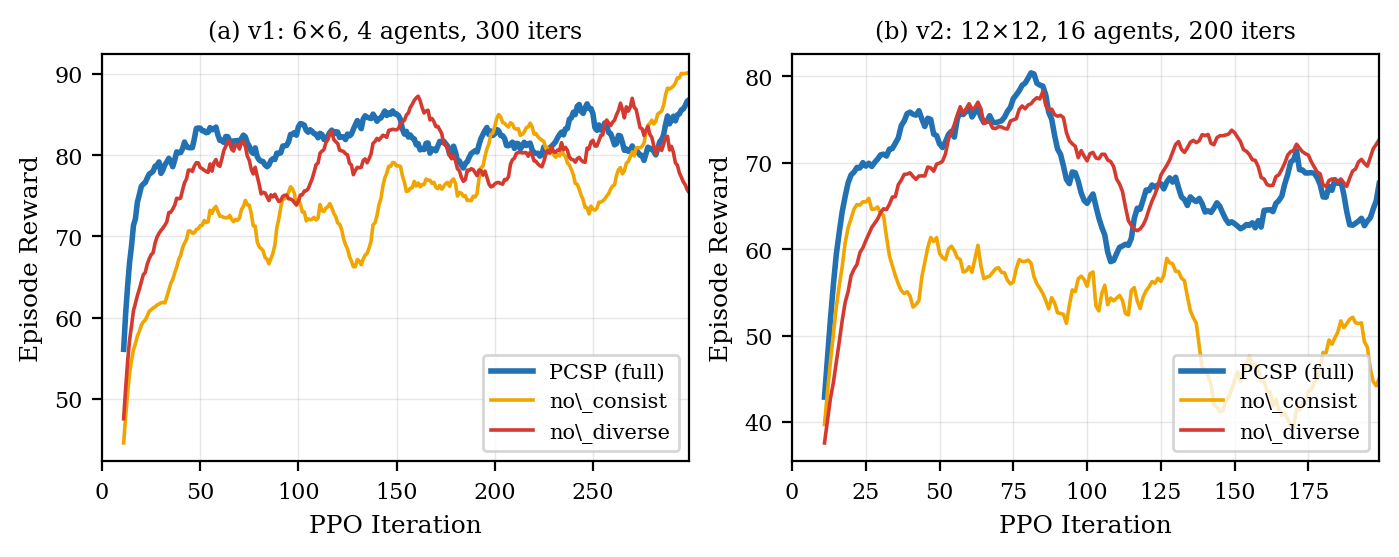}
  \caption{%
    \textbf{Training reward curves at two scales.}
    (a) v1 (6$\times$6, 4 agents, 300 iters).
    (b) v2 (12$\times$12, 16 agents, 200 iters).
    Removing the consistency loss (no\_consist) preserves or even slightly
    increases reward but collapses zero-shot persona traceability across all
    scales (Tables~\ref{tab:results_v1}--\ref{tab:results_v3}).
    Removing the diversity loss (no\_diverse) causes KL collapse at v1 and v2
    ($0.39$ and $0.48$) and a v1 reward drop, but its effect on v3 zero-shot
    accuracy is within the Wilson CI of full (\S\ref{sec:evidence}).
  }
  \label{fig:learning}
\end{figure*}

Tables~\ref{tab:results_v3}, \ref{tab:results_v1}, and~\ref{tab:results_v2}
show key metrics for each setting. Figure~\ref{fig:learning} shows training
reward curves at v1/v2.

\begin{table}[t]
  \centering
  \small
  \caption{%
    \textbf{v3 results} (6$\times$6, 4 agents, 300 personas, 20-action
    ontology; 240 train / 60 zero-shot held-out occupations).
    ZS Acc: zero-shot $k$-NN accuracy on 60 unseen-occupation personas
    (random $\approx\!1.7\%$); brackets are Wilson 95\% CIs.
    Coh.: trajectory coherence ratio (intra/inter persona cosine similarity).}
  \label{tab:results_v3}
  \setlength{\tabcolsep}{3pt}
  \begin{tabular}{lrrr}
    \toprule
    Model & Reward $\uparrow$ & ZS Acc $\uparrow$ & Coh.\ $\uparrow$ \\
    \midrule
    \PCSP (full)                 & 104.1 & 0.170 \scriptsize{[.13,.22]}              & 2.06 \\
    \PCSP (no\_consist)          & 118.4 & 0.017* \scriptsize{[.01,.04]}             & 1.07 \\
    \PCSP (no\_diverse)          & 122.1 & 0.160 \scriptsize{[.12,.21]}              & 2.05 \\
    \textbf{\PCSP (concat)}      & 107.0 & \textbf{0.283} \scriptsize{[.24,.34]}     & \textbf{7.60} \\
    B1 No-Persona                & 100.3 & ---                                        & --- \\
    B3 SBERT                     & 106.9 & ---                                        & --- \\
    \bottomrule
  \end{tabular}\\[2pt]
  {\footnotesize * Near-random (failure mode).}
\end{table}


\begin{table}[t]
  \centering
  \small
  \caption{%
    \textbf{Conditioning architecture $\times$ OOD split (v3).}
    Zero-shot $k$-NN accuracy on three compositional held-out splits, evaluated
    against the same trained PCSP checkpoint family (random $\approx\!1.7\%$).
    Wilson 95\% CIs in brackets; \textbf{bold} marks the higher point estimate
    per row, and \dag\ marks rows where Wilson intervals overlap.}
  \label{tab:ood_splits}
  \setlength{\tabcolsep}{3.5pt}
  \begin{tabular}{lcc}
    \toprule
    OOD split & FiLM & Concat \\
    \midrule
    Unseen occupations
      & 0.170 \scriptsize{[.13,.22]}
      & \textbf{0.283} \scriptsize{[.24,.34]} \\
    Unseen archetypes
      & \textbf{0.203} \scriptsize{[.16,.25]}
      & 0.103 \scriptsize{[.07,.14]} \\
    Unseen occ.\,$\times$\,archetype\,\dag
      & \textbf{0.203} \scriptsize{[.16,.25]}
      & 0.170 \scriptsize{[.13,.22]} \\
    \bottomrule
  \end{tabular}\\[2pt]
  {\footnotesize \dag Wilson intervals overlap; gap not significant.
   Architecture preference flips with the OOD axis: concat generalizes more
   strongly when the occupation is novel, FiLM when the archetype is novel.}
\end{table}

\subsection{Key Observations}

Across all three experimental settings, the ablation evidence converges on a
single load-bearing component---the InfoNCE consistency loss---while the
choice of conditioning architecture proves to be secondary.

\textbf{The consistency loss is the load-bearing component.}
Removing the InfoNCE consistency term collapses zero-shot persona
identification to chance in every setting we have tested:
v1 drops from 19.3\% to 1.7\% (random), v2 drops from 2.3\% to exactly 0\%
against a 1.0\% baseline, and v3 drops from 17.0\% to 1.7\% (Wilson CI
$[.01, .04]$).
The reward axis hides this failure entirely. At v1 the no\_consist ablation
even slightly \emph{increases} reward (84.3 vs.\ 83.4); at v3 the gap is
sharper still (118.4 vs.\ 104.1, +14 reward) while accuracy collapses by 15.3
percentage points.
Without the consistency objective, the policy chases reward through
persona-agnostic strategies whose trajectories are no longer traceable back
to the conditioning persona, regardless of how the persona signal is injected
into the network.

\textbf{The conditioning architecture is secondary, and conditional on the
type of OOD shift.}
We initially conjectured that FiLM's layer-wise gating would dominate input
concatenation, but the three v3 compositional splits in
Table~\ref{tab:ood_splits} show no universal winner. Concat dominates when
the held-out axis is occupation; FiLM dominates when the held-out axis is
Big-Five archetype; on the joint occ.\,$\times$\,archetype split the two are
statistically indistinguishable. Thus the conditioning mechanism is a
generalization-gap knob whose best setting depends on \emph{which} dimension
of the persona is novel at test time. The consistency loss, in contrast, is
the only component that universally fails across split families when removed.

\textbf{Diversity loss prevents behavioral collapse at v1/v2 but its effect
weakens on the v3 zero-shot accuracy axis.}
Removing the KL diversity term reduces mean policy KL by an order of
magnitude at v1 (5.87 to 0.39, $15\times$) and v2 (5.40 to 0.48, $11\times$),
matching the original behavioral-collapse story.
At v3 zero-shot, however, diversity-loss removal lands within the Wilson CI
of full (16.0\% vs.\ 17.0\%): the policy still achieves persona-recoverable
behavior on the accuracy axis without the explicit KL term.
We report diversity as a useful regularizer for KL-style behavioral spread
rather than as a co-equal contribution to zero-shot identification.

\textbf{Spearman $\rho$ is preserved across scales.}
\PCSP (full) achieves $\rho\!=\!0.728$ in v1 and $\rho\!=\!0.725$ in v2---a
difference of 0.003 despite a $4\times$ increase in grid area, $4\times$ more
agents, and $67\%$ more personas. This suggests that semantic-behavioral
alignment is a stable property of the design, not an artefact of the
minimal environment.

\subsection{Coarse-Trace Human Pilot}
\label{sec:humanpilot}

To test whether the policy's persona signal is visible to human readers, we ran
a 30-participant Google Forms pilot using the coarse Korean 2AFC survey. Each
item showed a PCSP-v3 trajectory as a sequence of coarse action labels and asked
participants to choose which of two persona descriptions generated it. Because
the form preserved only item-level A/B selection ratios, we report aggregate
forced-choice accuracy rather than participant-level variance, confidence, or
response-time analyses.

Participants selected the correct persona in 612 of 900 aggregate judgments
($68.0\%$; pooled Wilson 95\% CI $[64.9, 71.0]$), above the 50\% chance
baseline. Item-level results were mixed: 15 of 30 items were strongly readable
($\geq 80\%$ correct), 2 were moderately readable ($70$--$79\%$), 8 were
ambiguous ($40$--$69\%$), and 5 were misleading ($<40\%$). Thus coarse traces
do expose some persona-conditioned behavior, but they leave a substantial
minority of items ambiguous or inverted. We interpret this as evidence for the
observability bottleneck motivating richer trajectory rendering, not as a
completed rich-versus-coarse human comparison.

\subsection{Qualitative Case Study: Designer-Authored Personas}
\label{sec:qualcase}

To test whether \PCSP can be controlled by persona descriptions outside the
synthetic Big-Five/occupation templates, we constructed 50 designer-authored
personas drawn from five sources at varying distance from the training
distribution: \textit{The Sims 3} trait combinations~\cite{sims3traits}
(\eg Workaholic + Ambitious + Perfectionist; $n\!=\!10$),
\textit{Animal Crossing} villager personality
categories~\cite{nookipediaVillager} (\eg Lazy, Jock, Normal; $n\!=\!12$),
\textit{Stardew Valley} NPC archetypes (\eg the reclusive coder, the
off-grid hermit; $n\!=\!10$), \textit{Persona}-series confidant descriptions
(\eg the perfectionist student leader, the hikikomori hacker; $n\!=\!6$),
and original designer briefs authored from scratch from Big-Five composites
(\eg burned-out doctor, sleep-deprived startup founder; $n\!=\!12$). Each
source concept was rewritten as a 2--3 sentence natural-language NPC
description and assigned a plausible occupation.
All 50 designer-authored persona texts are written in \emph{English}, while
the 240 training personas are Korean; the case study therefore additionally
serves as a fully cross-lingual robustness probe of the frozen multilingual
Qwen3 embedding and the learned low-rank projection. No retraining was
performed: each text was encoded with the same Qwen3 embedding pipeline,
passed through the learned low-rank projection, and rolled out for five v3
episodes.

\textbf{Outcome and failure taxonomy.}
Each rollout is classified by overlap between its top-3 actions and the
authored preferred-action signature. Outcomes are \emph{success} (overlap
$\geq\!2$), \emph{partial} ($=\!1$), or \emph{failure} ($=\!0$). Failures are
further tagged by mechanism: \textbf{F1} ontology gap (authored behavior is
not expressible in the 20-action space, \eg ``protect,'' ``nurture''),
\textbf{F2} style-reward conflict (top-3 mean style-cosine with the persona
Big-Five vector is negative), \textbf{F3} embedding occupational bias (rollout
follows the nearest train-occupation neighbor rather than the authored
personality), \textbf{F4} trait collision (a declared multi-trait composite
collapses to a single trait), and \textbf{F5} residual (zero overlap with
none of the above triggers).

\begin{table}[t]
  \centering
  \small
  \caption{%
    \textbf{Designer-authored persona case study (50 English personas, 5
    sources, Korean-trained policy).}
    Outcome breakdown from five v3 rollouts per persona; failure-mode column
    lists the diagnoses present in that source's failure cases.
    Mean cosine similarity to the nearest (Korean) training persona is 0.590
    (range 0.478--0.686).}
  \label{tab:designer_personas}
  \setlength{\tabcolsep}{3.0pt}
  \begin{tabular}{lrrrrl}
    \toprule
    Source & $n$ & Succ. & Part. & Fail. & Failure modes \\
    \midrule
    The Sims 3              & 10 & 4 & 5 & 1 & F1 \\
    Animal Crossing         & 12 & 4 & 5 & 3 & F1, F2, F5 \\
    Stardew Valley          & 10 & 5 & 4 & 1 & F1 \\
    Persona series          &  6 & 4 & 2 & 0 & --- \\
    Original brief          & 12 & 5 & 5 & 2 & F2, F4 \\
    \midrule
    \textbf{Total}          & \textbf{50} & \textbf{22} & \textbf{21} & \textbf{7} &
      \multicolumn{1}{r}{\textbf{86\% non-failure}} \\
    \bottomrule
  \end{tabular}
\end{table}

\begin{figure*}[t]
  \centering
  \includegraphics[width=0.98\textwidth]{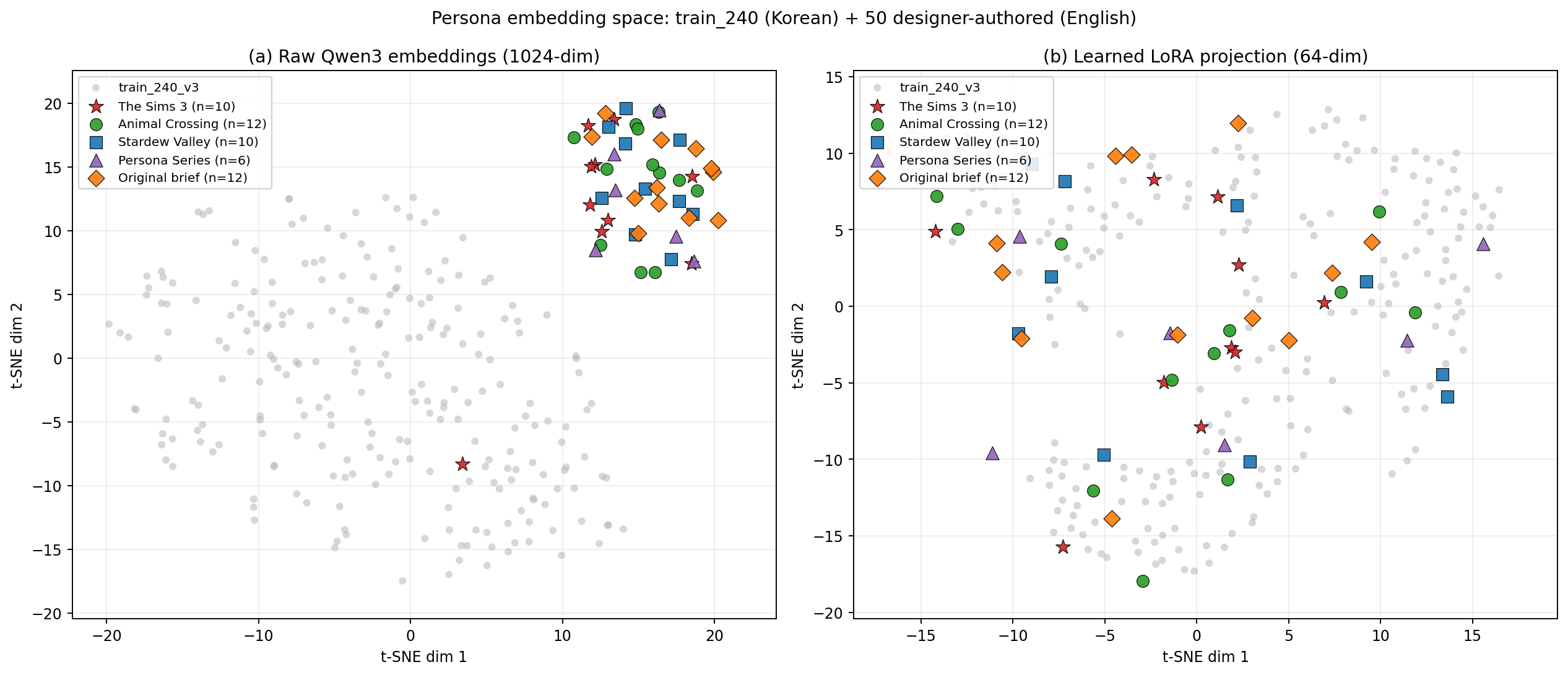}
  \caption{%
    \textbf{Designer-authored personas in embedding space: raw vs.\ learned
    projection.}
    t-SNE of 240 Korean training persona embeddings (grey) plus 50 English
    designer-authored personas, color-coded by source.
    \textbf{(a)} In the raw 1024-dim Qwen3 embedding space, the English
    designer personas form a language-shifted cluster well outside the Korean
    training distribution: their mean nearest-neighbor distance to any
    training persona is $12.9\times$ the mean train--train nearest-neighbor
    distance.
    \textbf{(b)} After the learned 64-dim low-rank persona projection used by the policy,
    the same designer personas are fully intermixed with training personas
    across all five sources (ratio drops to $1.47\times$).
    The projection is what makes the case-study setting effectively
    in-distribution for the trained policy, and is the load-bearing piece of
    the cross-lingual robustness reported in \S\ref{sec:qualcase}.}
  \label{fig:designer_tsne}
\end{figure*}

Three patterns are worth highlighting. First, 43 of 50 personas
(86\%) produce at least one top-3 action matching the authored intent, and
$22$ (44\%) match at least two---substantially above the chance baseline of
matching 2 of 3 from a 20-action set ($\approx\!9\%$). This holds even though
every designer-authored text is in English while the trained policy has only
ever seen Korean persona descriptions, providing direct evidence that the
multilingual Qwen3 backbone plus the learned low-rank projection yields a
language-agnostic persona representation. Figure~\ref{fig:designer_tsne}
makes the role of the projection explicit: in the raw Qwen3 space the English
designer personas form a clearly separated cluster ($12.9\times$ the
train--train nearest-neighbor distance), but after the learned low-rank persona projection
that the policy actually consumes (the low-rank persona projection) they are
fully intermixed with the Korean training personas ($1.47\times$). The mean cosine similarity between each
English persona and its nearest Korean training neighbor drops only modestly
(0.608 with the Korean-text variant of the same designer set vs.\ 0.590 here),
and per-source success rates are stable or higher across the language shift.

Second, source-stratified results expose a meaningful gradient: Stardew Valley
NPC archetypes generalize best (50\% strong / 40\% partial / 10\% failure),
The Sims 3 trait combinations and original Big-Five briefs sit at a similar
level (40\% / 50\% / 10\% and 42\% / 42\% / 17\%), and the Persona-series
confidants never fail outright (67\% / 33\% / 0\%). Animal Crossing personality
categories remain the hardest source (25\% failure), concentrated in F1, F2,
and F5 tags. This is consistent with the AC categories being defined almost
entirely by stylistic and affective markers (\eg Cranky, Snooty, Sisterly)
that the v3 ontology cannot distinguish at the action level.

Third, the dominant failure mechanism is ontology-limited rather than
embedding-limited. Of seven failures, five (F1$\times$3, F4, F5) involve
composite or stylistic concepts the 20-action ontology cannot distinguish:
the \emph{normal villager} (hygiene/care behavior, F1),
\emph{family-oriented caretaker} (protective family role, F1),
\emph{animal clinic worker} (gentle care semantics, F1),
\emph{sleep-deprived startup founder} (multi-trait collision, F4), and the
\emph{jock villager} (fitness-coded social action collapses into the generic
\texttt{socialize\_initiate} bucket, F5). Only two failures are squarely
policy-side: the \emph{cranky villager} and the \emph{restless rideshare
driver} (both F2, the per-action style reward outranks the persona-conditioned
intent). Notably, the F3 embedding-occupational-bias failure that appeared
in the Korean-text variant disappears here---with English designer text and
Korean training neighbors, the embedding can no longer latch onto a shared
occupation token. This mirrors the coarse-action observability problem
discussed in \S\ref{sec:limits} and is the cleanest argument in this work
that the next environment iteration should expand the social/stylistic axis
of the action space rather than the embedding or training pipeline.

\subsection{Cross-Scale Replication}

\begin{figure}[!htbp]
  \centering
  \includegraphics[width=\linewidth]{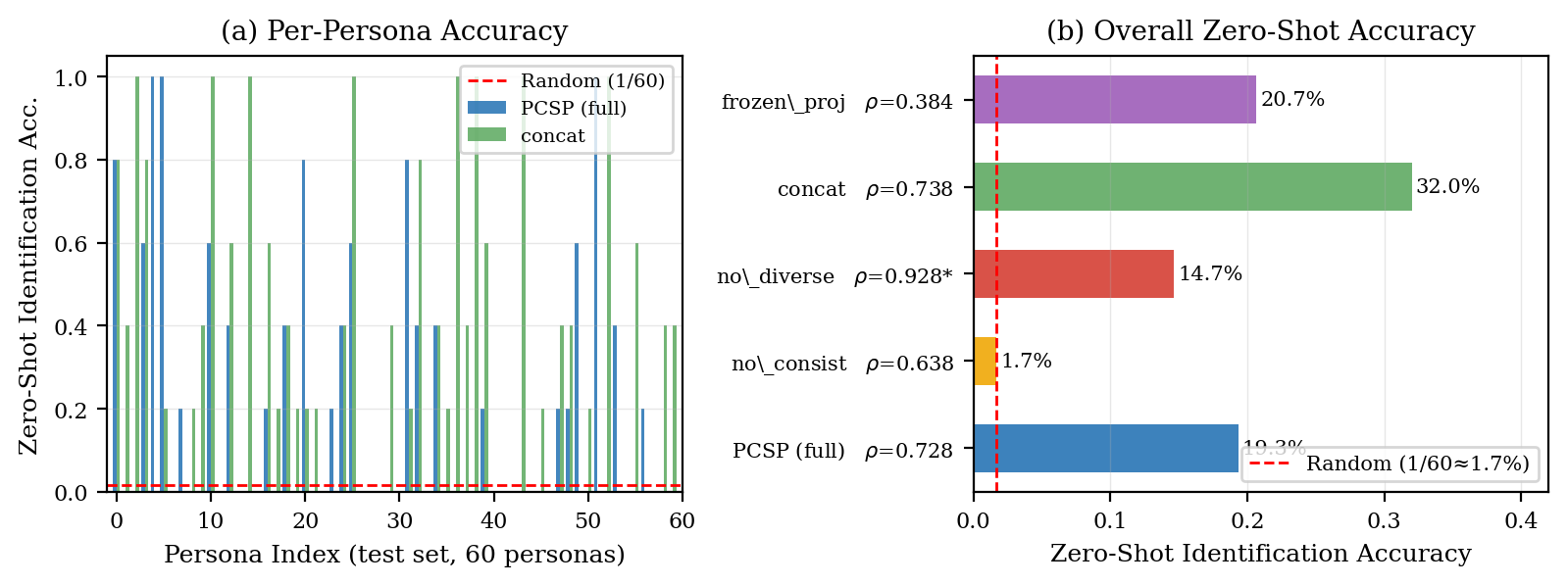}
  \caption{%
    \textbf{Zero-shot generalization on 60 unseen personas (v1).}
    \PCSP (full) achieves 19.3\% trajectory-to-persona identification;
    the red dashed line marks random chance (1.7\%).
    Removing the consistency loss collapses accuracy to 1.7\%.
    The v2 experiment (100 unseen personas) replicates this qualitative pattern.
  }
  \label{fig:zeroshot}
\end{figure}

\begin{figure}[!htbp]
  \centering
  \includegraphics[width=\linewidth]{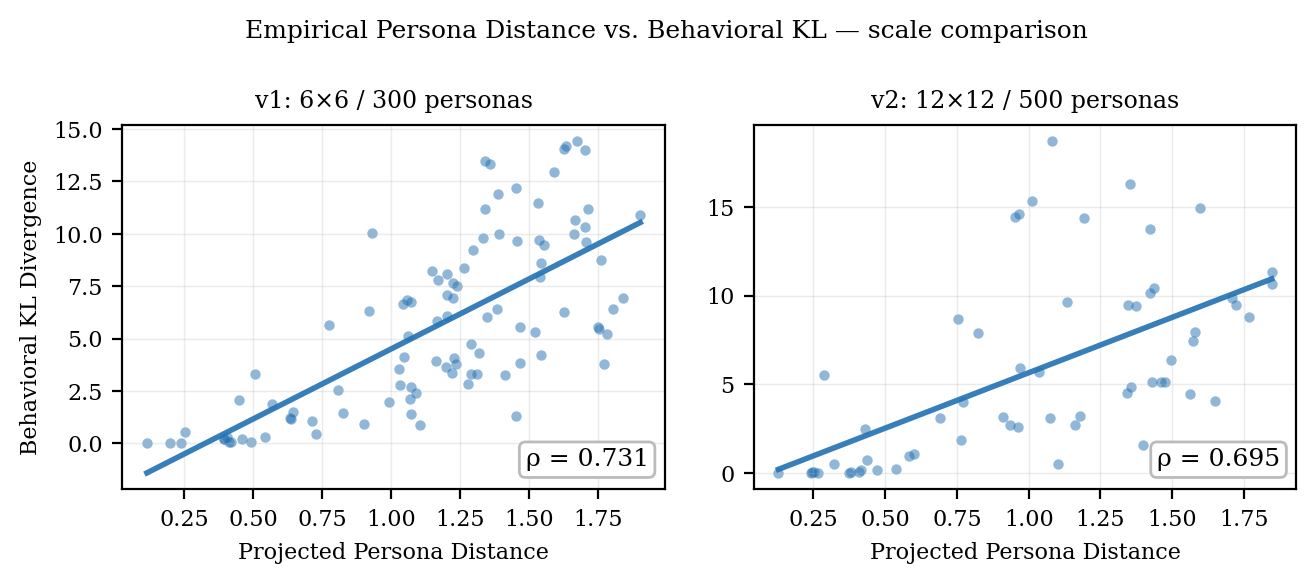}
  \caption{%
    \textbf{Empirical projected persona distance vs.\ behavioral KL divergence
    at two scales (PCSP full).}
    Each point is a recomputed persona pair from the trained policy checkpoints
    rather than a synthetic reconstruction from summary statistics. Left: v1
    sampled-pair $\rho\!=\!0.731$ (100 pairs, 200 states). Right: v2
    sampled-pair $\rho\!=\!0.695$ (60 pairs, 100 states). The monotone
    alignment between persona space and behavior space is preserved when moving
    to a 4$\times$ larger grid with 4$\times$ more agents and 67\% more personas.
  }
  \label{fig:kl}
\end{figure}

The absolute task reward and zero-shot accuracy are lower in v2, as expected for
a substantially harder environment with more test personas.
The same empirical structure is observed across environment scales: the InfoNCE
consistency objective is required for zero-shot persona traceability, and the
semantic-behavioral alignment property ($\rho \approx 0.73$) is preserved.
Figure~\ref{fig:zeroshot} shows the zero-shot distribution for v1;
the v2 pattern is qualitatively similar with narrower per-persona bars.
The aggregate evaluation gives nearly identical semantic-behavioral alignment
across environments ($\rho\!=\!0.728$ vs.\ $0.725$). Figure~\ref{fig:kl}
visualizes an independent empirical sample of persona pairs from the same
checkpoints, again showing strong monotone alignment at both scales.


\section{Layer 2 Results: Cross-Substrate Generalization (Melting Pot)}
\label{sec:meltingpot}

\PCSPD{} isolates persona-conditioned control in a deliberately small grid
world; Layer 2 tests whether the method survives a different observation
geometry, action ontology, and social-dilemma structure. Cooperative
multi-agent benchmarks such as Hanabi~\cite{bard2020hanabi} and centralized-critic
algorithms~\cite{foerster2018coma,lowe2017maddpg} target related challenges of
joint policy learning, but do not address per-agent persona conditioning. We apply \PCSP{}
unchanged---no algorithmic modifications beyond the CNN front-end required
by RGB inputs---to three \textit{Melting Pot
2.4.0}~\cite{leibo2021meltingpot} substrates spanning three distinct
social-dilemma categories: \texttt{commons\_harvest\_\_open} (commons-pool
resource), \texttt{clean\_up} (public-good provisioning), and
\texttt{prisoners\_dilemma\_in\_the\_matrix\_\_repeated} (dyadic
matrix-game with distinct action ontology and 40$\times$40 RGB). The
substrates were chosen \emph{before} training and were not tuned against.
We evaluate persona identifiability with the same protocol as Layer 1 and
re-run the InfoNCE ablation in every substrate.

\noindent\textbf{Substrate.}
\texttt{commons\_harvest\_\_open} is a 7-player common-pool resource substrate with
$88\!\times\!88\!\times\!3$ RGB observations and an 8-action ontology
(\texttt{NOOP}, \texttt{FORWARD}, \texttt{BACKWARD}, \texttt{STEP\_LEFT/RIGHT},
\texttt{TURN\_LEFT/RIGHT}, \texttt{FIRE\_ZAP}). Two MP-specific additions are
required and otherwise the algorithm matches \S\ref{sec:method} byte-for-byte:
(i) a 3-layer Nature-style CNN
($32\!\times\!8/4 \to 64\!\times\!4/2 \to 64\!\times\!3/1$) maps the RGB
observation to a 256-d feature that feeds the unchanged FiLM trunk, and (ii) a
small persona-agnostic shaping term $\tilde r_i = r_i + 0.1\cdot\bar r$
(team-mean potential) is added to escape the well-known apple-depletion trap on
which vanilla PPO collapses to zero return. All other components---frozen
Qwen3-0.6B-Embedding, rank-$16$ low-rank persona projection $1024\!\to\!64$, FiLM
trunk 256-256-128, 2-layer GRU trajectory encoder, full-batch InfoNCE at
$T\!=\!0.07$ and $\lambda_{\text{IC}}\!=\!0.5$, KL diversity at
$\lambda_2\!=\!0.1$ over $8$ personas $\times\,32$ states, and PPO
($\gamma\!=\!0.99$, GAE-$\lambda\!=\!0.95$, clip $0.2$, entropy $0.01$)---are
identical to \S\ref{sec:method}. Personas are the same 12-persona Qwen3
corpus as the companion technical report~\cite{phase5report}, split
$10$ train / $2$ held-out; this section reports in-distribution metrics on the
$10$ train personas at $1\,\text{M}$ environment steps per seed. Under the
shaping, agents earn modest non-zero return throughout training (mean per-step
reward $\approx\!0.010$ across all 5 seeds, second half of training), so the
substrate is engaged rather than collapsed; commons\_harvest is well known to be
hard to fully solve with vanilla PPO and we do not attempt to do so here.

We measure (i) mean pairwise action-KL between persona-conditioned policies on
$64$ random rollout states, averaged over the $\binom{10}{2}\!=\!45$ persona pairs,
and (ii) trajectory$\to$persona retrieval top-3 accuracy with the trained GRU
encoder over the $10$-train vocabulary (chance $3/10\!=\!0.30$). Bootstrap 95\%
CIs use $10\,000$ resamples over seeds.

\begin{table*}[t]
  \centering
  \small
  \caption{%
    \textbf{Melting Pot multi-substrate validation (Tab.~3).} Final-checkpoint
    metrics at $1\,\text{M}$ environment steps per seed across three substrates.
    Mean pairwise action-KL (10 train personas, 45 pairs, 64 random states)
    and trajectory$\to$persona retrieval on the 10-train vocabulary
    (chance top-1 $\!=\!0.10$, top-3 $\!=\!0.30$).
    The \emph{no-InfoNCE} ablation collapses retrieval to chance in every
    substrate while leaving (or inflating) pairwise KL---the consistency loss
    is load-bearing across all three social-dilemma categories.
    Numbers are mean$\pm$std over seeds; \texttt{commons\_harvest\_\_open}
    full has 5 seeds and a single ablation seed for backward compatibility,
    the two new substrates have 3 seeds per condition.}
  \label{tab:mp_multi}
  \setlength{\tabcolsep}{5pt}
  \begin{tabularx}{\textwidth}{>{\raggedright\arraybackslash}X l c c c r}
    \toprule
    Substrate (category) & Config & seeds & top-1 & top-3 & pair KL \\
    \midrule
    \texttt{commons\_harvest\_\_open}  & Full \PCSP                & 5 & $0.564\!\pm\!0.209$ & $0.936\!\pm\!0.073$ & $2.11\!\pm\!0.70$ \\
    (commons-pool)                      & $-\!$InfoNCE              & 1 & $0.071$            & $0.250$            & $5.82$            \\
    \midrule
    \texttt{clean\_up}                  & Full \PCSP                & 3 & $\mathbf{0.690\!\pm\!0.206}$ & $\mathbf{1.000\!\pm\!0.000}$ & $1.01\!\pm\!0.29$ \\
    (public-good)                       & $-\!$InfoNCE              & 3 & $0.095\!\pm\!0.041$ & $0.226\!\pm\!0.055$ & $2.91\!\pm\!1.12$ \\
    \midrule
    \texttt{prisoners\_dilemma\_\dots\_repeated} & Full \PCSP       & 3 & $\mathbf{0.625\!\pm\!0.217}$ & $\mathbf{1.000\!\pm\!0.000}$ & $2.87\!\pm\!0.74$ \\
    (dyadic matrix-game)               & $-\!$InfoNCE              & 3 & $0.125\!\pm\!0.125$ & $0.167\!\pm\!0.144$ & $4.54\!\pm\!0.24$ \\
    \midrule
    chance (10 train personas)         & ---                       & --- & $0.10$ & $0.30$ & $0$ \\
    \bottomrule
  \end{tabularx}
\end{table*}

\begin{figure}[t]
  \centering
  \includegraphics[width=\linewidth]{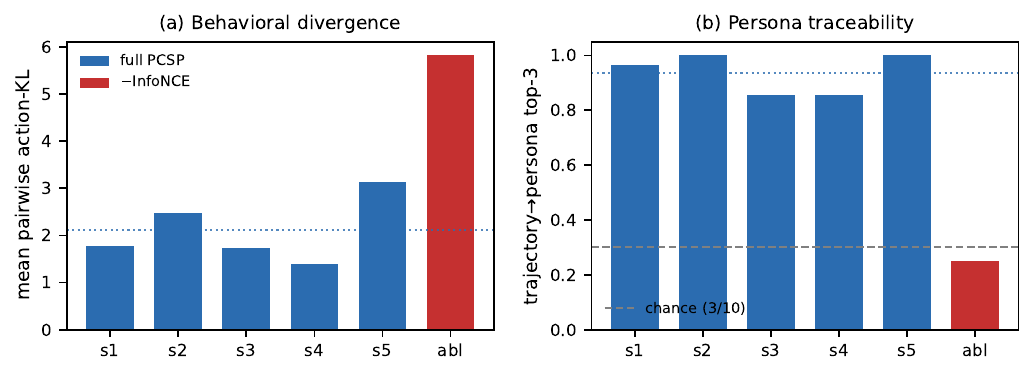}
  \caption{%
    \textbf{Melting Pot single-substrate detail on \texttt{commons\_harvest\_\_open}.}
    Per-seed bars (blue: full \PCSP, 5 seeds; red: $-$InfoNCE ablation, 1 seed).
    Dotted blue line: 5-seed mean. (a) Mean pairwise action-KL across the
    $\binom{10}{2}\!=\!45$ persona pairs. (b) In-distribution
    trajectory$\to$persona top-3 retrieval on the 10-train vocabulary
    (dashed grey: chance $3/10$). The ablation reaches the highest KL but the
    lowest top-3 (below chance): policies are different yet trajectories no
    longer encode \emph{which} persona produced them. The two additional
    substrates in Tab.~\ref{tab:mp_multi} replicate this pattern.}
  \label{fig:mp_clean}
\end{figure}

\paragraph{Takeaways.}
The \S\ref{sec:method} method---applied with no algorithmic modifications beyond
the CNN observation encoder required by RGB inputs and (on
\texttt{commons\_harvest\_\_open}) the persona-agnostic team-reward potential
that escapes the apple-depletion trap---produces persona-distinguishable
behavior across three Melting Pot substrates that differ in social structure
(commons-pool / public-good / dyadic-matrix), observation geometry, and
agent count (Table~\ref{tab:mp_multi}). The no-InfoNCE ablation reproduces
the v1--v3 finding in \emph{every} substrate: pairwise KL is held or inflated
($+38\%$ on commons\_harvest, $+1.9\times$ on clean\_up, $+58\%$ on
prisoners\_dilemma) while trajectory$\to$persona retrieval collapses to
chance ($0.10$ top-1 in all three new substrates vs.\ $0.56$--$0.69$ for the
full objective). The consistency loss is what makes the policy
\emph{traceable to its conditioning}, not merely diverse, and this property
is invariant across the three social-dilemma categories tested.

\subsection{Held-out persona recovery and cross-substrate transfer}
\label{sec:mp_transfer}

The 12-persona corpus is split $10$ train / $2$ held-out
(\texttt{fast\_mover}, \texttt{spinner}). The substrate-internal results
above use the 10-train vocabulary. Here we ask two additional Layer-2
questions that the in-distribution table cannot: (i) does the InfoNCE-trained
manifold extend to the held-out persona embeddings; and (ii) does the
manifold learned on one substrate transfer to trajectories collected on a
different substrate.

\noindent\textbf{Protocol.} For each Layer-2 checkpoint we run a $256$-step
rollout in the substrate's environment, with every agent independently
conditioned on a persona embedding drawn from the full 12-vocab. Per-agent
trajectories are encoded by the trained $\text{GRU}$ trajectory encoder and
compared against the 12 \emph{projected} persona embeddings via cosine
similarity; chance top-1 is $1/12\!\approx\!0.083$ and chance top-3 is
$0.25$. Cross-substrate transfer between
\texttt{commons\_harvest\_\_open} (CH) and \texttt{clean\_up} (CU) reuses
the source substrate's persona projection and trajectory encoder
(zero-padded on the GRU input weight from $8$ to $9$ action one-hots when
needed; no other parameter is touched) and applies them to trajectories
collected by the target substrate's policy in the target environment.
Observation shapes match across CH and CU ($88\!\times\!88\!\times\!3$);
\texttt{prisoners\_dilemma} is excluded from cross-substrate transfer
because of its $40\!\times\!40$ RGB.

\begin{table}[t]
  \centering
  \small
  \caption{%
    \textbf{Held-out-vocab and cross-substrate transfer (Tab.~4).}
    \emph{Top:} retrieval against the full 12-persona vocabulary
    (chance top-1 $=0.083$, top-3 $=0.25$); seeds match Tab.~\ref{tab:mp_multi}.
    \emph{Bottom:} cross-substrate transfer over CH and CU full PCSP
    checkpoints (3 seed pairs, chance top-1 $=0.10$, top-3 $=0.30$).
    Mean$\pm$std over seeds.}
  \label{tab:mp_transfer}
  \setlength{\tabcolsep}{4pt}
  \begin{tabular}{l l c c c}
    \toprule
    \multicolumn{2}{l}{Setting} & seeds & top-1 & top-3 \\
    \midrule
    \multicolumn{5}{l}{\emph{Held-out vocabulary (12-persona)}} \\
    CH                & Full \PCSP            & 5 & $0.286\!\pm\!0.120$ & $0.657\!\pm\!0.153$ \\
                      & $-\!$InfoNCE          & 1 & $0.000$            & $0.357$            \\
    CU                & Full \PCSP            & 3 & $\mathbf{0.405\!\pm\!0.168}$ & $\mathbf{0.810\!\pm\!0.089}$ \\
                      & $-\!$InfoNCE          & 3 & $0.071\!\pm\!0.000$ & $0.262\!\pm\!0.089$ \\
    PD                & Full \PCSP            & 3 & $\mathbf{0.333\!\pm\!0.068}$ & $\mathbf{0.889\!\pm\!0.104}$ \\
                      & $-\!$InfoNCE          & 3 & $0.028\!\pm\!0.039$ & $0.167\!\pm\!0.068$ \\
    \midrule
    \multicolumn{5}{l}{\emph{Cross-substrate transfer (CH$\leftrightarrow$CU, 10-train vocab)}} \\
    \multicolumn{2}{l}{CH $\to$ CU}                            & 3 & $0.060\!\pm\!0.034$ & $0.417\!\pm\!0.034$ \\
    \multicolumn{2}{l}{CU $\to$ CH}                            & 3 & $\mathbf{0.179\!\pm\!0.058}$ & $\mathbf{0.429\!\pm\!0.077}$ \\
    \bottomrule
  \end{tabular}
\end{table}

\noindent\textbf{Results.}
Three things stand out (Tab.~\ref{tab:mp_transfer}). \emph{(i) Full PCSP
generalises to the 12-vocab.} Full-objective retrieval over the
12-persona vocabulary remains $3.4$--$4.9\times$ chance for top-1 and
$2.6$--$3.6\times$ chance for top-3 in every substrate, and the no-InfoNCE
ablation collapses to chance in every substrate (top-1 $\le 0.071$). The
mechanism is robust to vocabulary expansion. \emph{(ii) Held-out persona
recovery is the open problem.} Across all 11 full \PCSP{} runs, the held-out
top-1 is $0.000$: the two held-out personas (\texttt{fast\_mover},
\texttt{spinner}) never retrieve themselves. This reproduces the embedding-margin
condition documented in the companion technical report~\cite{phase5report}:
the held-out embeddings sit inside the convex hull of the train
embeddings and the trained projection does not separate them. We flag this
as the most direct remaining Layer-2 gap. \emph{(iii) Cross-substrate
transfer is asymmetric.} CU $\to$ CH yields top-1 $0.179\!\pm\!0.058$
($1.79\times$ chance) and top-3 $0.429\!\pm\!0.077$ ($1.43\times$ chance),
showing that the persona manifold learned on the richer \texttt{clean\_up}
substrate carries information that survives transplantation into the CH
environment. The reverse direction CH $\to$ CU is at-or-below chance for
top-1 ($0.060\!\pm\!0.034$) but recovers a non-trivial top-3 signal
($0.417\!\pm\!0.034$, $1.39\times$ chance), indicating partial transfer at
the rank-3 level. The asymmetry---a positive cross-task signal, but
direction-dependent---is consistent with the InfoNCE manifold's
substrate-specific shaping; building a substrate-invariant projection is a
clean follow-up.

\section{Layer 3 Results: Realtime Engine Deployment (UE5)}
\label{sec:ue5}

\noindent\textbf{Layer 3 framing.}
A persona-conditioned policy is only meaningful if it survives the
engineering pressure of a real game engine: asynchronous tick rates,
NavMesh contention, BT failure recovery, ONNX runtime constraints, and
shared world state. We deploy a \emph{frozen} Layer-1 (\PCSPD{}-v3)
checkpoint into Unreal Engine 5.5 as a hybrid intent stack
(architecture and PIE screenshots in App.~\ref{app:ue5_system},
Figs.~\ref{fig:ue5_system}--\ref{fig:ue5_debugger}): \PCSP{} selects
semantic intents (e.g.\ \texttt{EatQuick}, \texttt{FocusedWork},
\texttt{LeisureOutdoor}) and the Behavior Tree, Blackboard, EQS,
AIController, and NavMesh execute them. The policy is exported to ONNX
(20-action head, $33$-d obs, $64$-d persona projection) and run through
Epic's NNE/NNERuntimeORT plugin from a custom
\texttt{UPCSPPolicySubsystem}; persona embeddings are loaded once at
level start from the same \texttt{persona\_embeddings.json} produced by
the research pipeline, and the policy network itself is the unmodified
\PCSP{} checkpoint (no engine-side training).
With this stack we ask three questions that Layers 1 and 2 cannot
answer: (i) does the policy meet a realtime wall-budget at deployment
scale; (ii) does the InfoNCE finding survive engine-side contention;
(iii) do personas maintain identity over horizons far longer than the
training episode.

\textbf{Realtime envelope (summary).}
A single-workstation scaling sweep
($\{8,16,32,64,96,128\}$ agents $\times\;3$ seeds $\times\;630$\,s in
standalone \texttt{-game} mode; full protocol and per-setting numbers in
App.~\ref{app:ue5_scaling}, Table~\ref{tab:ue5_scaling}) establishes that
ONNX inference is \emph{not} the bottleneck (mean per-call latency
$183$--$202$\,\textmu s through $n{=}64$); frame time scales near-linearly
at $\sim\!0.27$\,ms/agent; and the hard ceiling is NavMesh \texttt{FindPath}
saturation, not policy inference (BT-abort rate $\leq\!0.2\%$ for $n{\leq}64$,
$4.7\%$ at $n{=}96$, $44.9\%$ at $n{=}128$). We therefore fix
$n\!=\!64$ as the operating point for every Layer-3 finding below.

\begin{table}[t]
  \centering
  \small
  \caption{%
    \textbf{Held-out persona transfer (UE5, 64 agents, identical map).}
    Train: personas $1$--$240$; Held-out: $60$ test personas (IDs
    $241$--$300$) covering four occupations absent from training.
    $\rho_{\text{intra}}$ is inter-persona action-histogram Spearman.}
  \label{tab:ue5_heldout}
  \setlength{\tabcolsep}{4pt}
  \begin{tabular}{lrrrrr}
    \toprule
    Personas & Min.\ & $n_{\text{int}}$ & Int/agent & Fail \% & $\rho_{\text{intra}}$ \\
    \midrule
    Train (1--240)   & 5.0  & 1{,}474 & 33.5 & 0.00 & 0.383 \\
    \textbf{Held-out (241--300)} & 9.75 & \textbf{2{,}792} & \textbf{43.6} & \textbf{0.04} & \textbf{0.368} \\
    \bottomrule
  \end{tabular}
\end{table}

\textbf{Held-out persona generalization in-engine.}
We ran a clean zero-shot evaluation with the $60$ held-out personas
(IDs 241--300) inside UE5 after matching environment capacity to the held-out
demand profile (Office, Hygiene; Table~\ref{tab:ue5_heldout}). Result: $2{,}792$ interactions in $9.75$\,min,
\textbf{0.04\% failure rate} ($1$ path-follow event of $2{,}792$),
$43.6$ interactions per agent (vs.\ $33.5$ on train personas), reward $960.4$
(vs.\ $730.6$), category coverage $9/10$ in both. The inter-persona action
dispersion is $\rho = 0.368$ on held-out personas vs.\ $0.383$ on train---the
engine-integrated policy is \emph{slightly more} persona-distinct on personas it
has never seen, matching the \PCSPD-v3 zero-shot finding qualitatively.

\begin{table}[t]
  \centering
  \small
  \caption{%
    \textbf{In-engine ablation at 64 agents, $\sim$5\,min per run}
    (\texttt{pcsp.PolicyMode} CVar, identical map and persona set).
    $n_{\text{int}}$: \texttt{interaction\_complete} events.
    Fail: \texttt{move\_failed} / total decisions.
    $\rho_{\text{intra}}$: mean Spearman over $\binom{64}{2}$ persona pairs
    on action histograms. Sym.\ KL is against the reference
    \texttt{HybridPCSP} run.}
  \label{tab:ue5_ablation}
  \setlength{\tabcolsep}{3pt}
  \begin{tabular}{lrrrrr}
    \toprule
    Mode & $n_{\text{int}}$ & Fail \% & Reward & $\rho_{\text{intra}}$ & Sym.\ KL \\
    \midrule
    \textbf{HybridPCSP}    & \textbf{2{,}077} & \textbf{0.0}  & \textbf{708.9} & \textbf{0.368} & --- \\
    BTOnly                 & 1{,}152 & 87.6 & 395.2 & 0.989 & ---$^\dagger$ \\
    HybridNoPersona        & 1{,}752 & 13.3 & 573.9 & 0.990 & 1.05 \\
    \bottomrule
  \end{tabular}\\[2pt]
  \raggedright\footnotesize
  $^\dagger$BTOnly emits zero logits by construction; KL against
  HybridPCSP would just measure distance from uniform.
\end{table}

\begin{figure}[t]
  \centering
  \includegraphics[width=\linewidth]{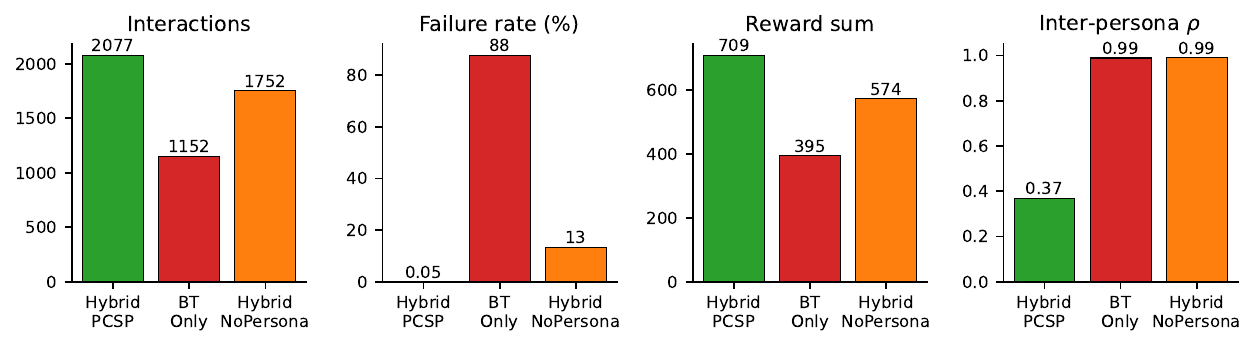}
  \caption{%
    \textbf{Phase 4 runtime ablation, 64 agents.} Zeroing the persona
    embedding (HybridNoPersona) collapses inter-persona action
    dispersion from $\rho\!=\!0.37$ to $\rho\!=\!0.99$. Replacing the
    ONNX policy with the BT-only needs heuristic (BTOnly) further
    halves throughput and reward as 64 agents synchronise on the
    single most-urgent need each tick.
  }
  \label{fig:ue5_ablation}
\end{figure}

\textbf{In-engine ablation.}
We re-ran the persona-conditioning ablation through a runtime CVar
(\texttt{pcsp.PolicyMode}) at 64-agent scale (5\,min each,
Table~\ref{tab:ue5_ablation}, Fig.~\ref{fig:ue5_ablation}):
full \PCSP achieved $2{,}077$ interactions / $0.0\%$ failure / reward $708.9$ /
action dispersion $\rho = 0.368$; zeroing the persona vector
(\texttt{HybridNoPersona}, the inference-time analogue of RL-only) raised
dispersion to $0.990$, increased failure to $13.3\%$, and yielded a symmetric KL
of $1.05$ against the full-PCSP policy distribution; bypassing ONNX entirely
(\texttt{BTOnly}, needs heuristic) collapsed to $1{,}152$ interactions / $87.6\%$
failure / dispersion $0.989$. Persona conditioning is load-bearing in the engine
under the same metric---action dispersion collapses by $2.7\times$ when the
persona signal is removed---reproducing the \S\ref{sec:evidence} pattern with no
algorithmic change. To calibrate the KL axis, two back-to-back PCSP runs on the
same map and persona set yielded a per-persona symmetric KL of $0.71$\,nats
(median $0.48$, Spearman $\rho = 0.65$), so the HybridNoPersona KL of $1.05$ is
already $1.5\times$ this paired-run noise floor.

\textbf{Consistency-loss replication in-engine.}
We additionally exported the v3 \texttt{no\_consist} checkpoint to ONNX and ran
paired 64-agent PIE sessions under identical \texttt{HybridPCSP} mode but with
the two different weight files. Full reached $3{,}110$ interactions in
$658$\,s with reward $1{,}079.8$ and inter-persona action $\rho = 0.379$;
\texttt{no\_consist} reached $4{,}005$ interactions in $681$\,s with reward
$1{,}423.5$ and $\rho = 0.312$. The two checkpoints choose meaningfully
different actions per persona at matched conditioning---mean Spearman
$\rho_{\text{match}} = 0.348$ and mean symmetric KL of $1.79$\,nats across the
$64$ personas---yet aggregate task reward is \emph{higher} for the checkpoint
without the consistency loss. This is the in-engine replication of the
\S\ref{sec:evidence} ``reward hides the failure'' result.

\subsection{Long-horizon behavioural persistence}
\label{sec:ue5_persistence}

\begin{figure}[t]
  \centering
  \includegraphics[width=\linewidth]{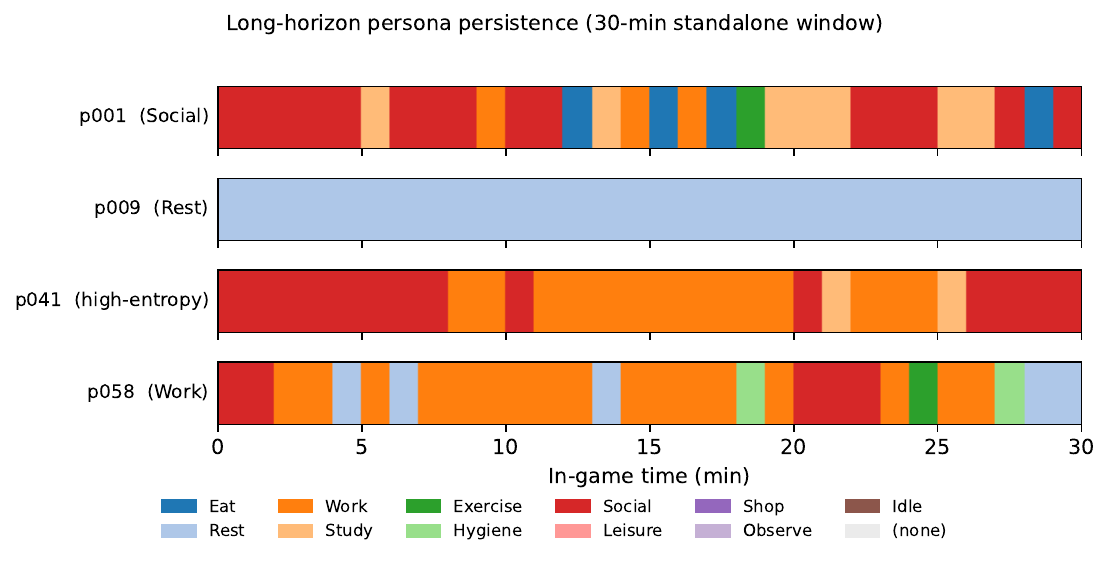}
  \caption{%
    \textbf{Long-horizon persona persistence (Layer~3).}
    Per-minute dominant intent category for four maximally-separated
    personas over $30$ in-game minutes in UE5 ($8$ agents, \emph{frozen}
    Layer-1 checkpoint, $0.5$\,s decision throttle, $0.0\%$ BT-abort).
    Each row is one persona; each column is a $1$-minute bin coloured by
    the modal intent category. \texttt{p009} stays in \texttt{Rest} for
    all $30$ bins (single run); \texttt{p058} holds \texttt{Work} as its
    modal axis in $17/30$ bins; \texttt{p001} is Social-dominant
    ($15/30$ bins) while cycling through five categories; \texttt{p041}
    splits evenly between \texttt{Social} and \texttt{Work}. No persona
    ever ``forgets'' its dominant axis over the $30$-minute window.}
  \label{fig:persona_persistence}
\end{figure}

\textbf{Do personas hold their routines past the training horizon?}
The Layer-1 training episode is $128$ environment steps (intent
decisions). To test whether persona identity persists over horizons far
longer than the policy ever saw at training time --- and under a
contention level low enough that the signal is not masked by
zone-capacity collisions (cf.\ \S\ref{sec:ue5_contention}) --- we ran a
dedicated low-density session: $8$ agents in standalone \texttt{-game}
for \texttt{pcsp.RunDurationSeconds}${=}1800$
(\texttt{ue/cnzoi/Saved/PCSP/Logs/20260520\_102022}, $1{,}794$\,s,
\textbf{$0.0\%$ BT-abort}). The eight agents are pinned to four personas
(two agents each) chosen \emph{a~priori} to maximise pairwise
category-distribution sym-KL on an independent clean reference session
(\texttt{20260518\_114841}; minimum pairwise sym-KL among the four
$=2.68$ nats): \texttt{p001} (Social-leaning), \texttt{p009}
(Rest-leaning), \texttt{p041} (high-entropy), and \texttt{p058}
(Work-leaning). Each agent issues $119$--$291$ sequential intent
decisions over the window ($0.9$--$2.3\times$ the $128$-step training
episode), spanning $30$ minutes of wall-clock realtime. For each persona
we bin every \texttt{decision} / \texttt{interaction\_complete} event
into $1$-minute windows and report the dominant intent category per bin
(Fig.~\ref{fig:persona_persistence}, derived by
\texttt{research/\allowbreak scripts/\allowbreak build\_\allowbreak persona\_\allowbreak persistence.py}; raw bin
sequence and per-bin category histograms in
\texttt{research/\allowbreak results/\allowbreak ue\_sessions/\allowbreak 20260520\_\allowbreak 102022/\allowbreak persona\_\allowbreak persistence.json}).

Three observations. (i) \emph{Persona-level stability holds past the
training horizon.} The top-category share over the $30$-minute window
is $1.00$ / $0.57$ / $0.50$ / $0.47$ for \texttt{p009} / \texttt{p058} /
\texttt{p001} / \texttt{p041} respectively---the persona ordering by
``focus'' is preserved end-to-end, and the most-focused persona
(\texttt{p009}, $291$ decisions/agent) stays on its preferred axis for
every single bin. (ii) \emph{Persistence is persona-specific, not flat.}
\texttt{p001} visits five distinct categories with $18$ inter-bin
transitions; \texttt{p009} executes a single $30$-bin \texttt{Rest} run.
The figure therefore shows persona persistence \emph{without} flattening
into ``every persona does its top thing forever''---which would be
indistinguishable from a learned task-side bias. (iii) \emph{The
mechanism is the conditioning, not a recurrent state.} The policy is
stateless feed-forward over the per-step observation; the only
persona-side memory is the frozen $64$-d projection. Behavioural
persistence over $30$ minutes of realtime is thus purely a property of
the persona embedding plus the InfoNCE-trained conditioning manifold,
not of any in-engine temporal smoothing.

\subsection{Failure analysis and contention: the $\rho$-drop as a Layer-3 finding}
\label{sec:ue5_contention}

\textbf{BT failure taxonomy.}
The hybrid stack distinguishes three engine-side failure modes, all emitted
as Behaviour-Tree aborts that the offline analyser bins by
\texttt{failure\_reason} (Table~\ref{tab:ue5_failures}). Across nine 64-agent
PIE sessions
(including both held-out and runtime ablation conditions),
\texttt{FindBestZone:AllOverCapacity}---every candidate zone for the chosen
intent already at capacity---accounts for the dominant share of aborts in
contention-heavy regimes (e.g.\ 87.6\% of decisions in the BTOnly run, where
$64$ agents synchronise on the most-urgent need each tick). Under the full
\PCSPD-conditioned stack with tuned capacities this collapses to
$<\!2$\% of decisions, with the residual being
\texttt{zone\_no\_free\_interaction\_point} (a chosen zone fills between
selection and arrival) and \texttt{path\_follow\_idle\_short} (a NavMesh
follower stalls below a movement threshold). Critically, the
\texttt{HybridNoPersona} ablation raises the abort rate from $0.0\%$ to
$13.3\%$ \emph{without changing the map}---the persona signal is what
desynchronises agents into different categories and keeps the contention
budget liveable.

\begin{figure}[t]
  \centering
  \includegraphics[width=\linewidth]{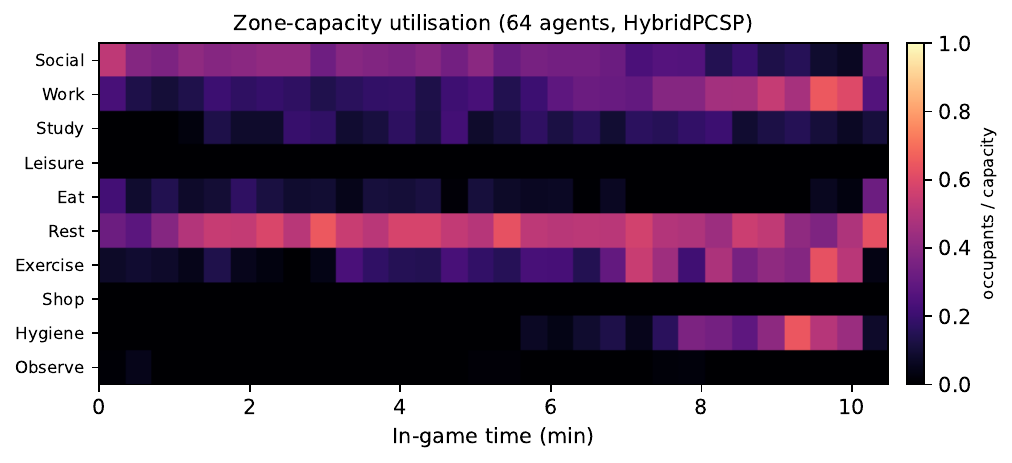}
  \caption{%
    \textbf{Zone-capacity utilisation over a $64$-agent HybridPCSP
    episode.} Per-zone occupants$/$capacity (rows) across $30$ equal
    time bins over the $\sim\!10.5$-min episode (columns). Rest, Work, Hygiene, and Exercise zones
    carry the load (peaks $\approx\!0.65$); the unevenness is the
    engine-side contention behind the $\rho$-drop.}
  \label{fig:ue5_contention}
\end{figure}

\begin{figure}[t]
  \centering
  \includegraphics[width=\linewidth]{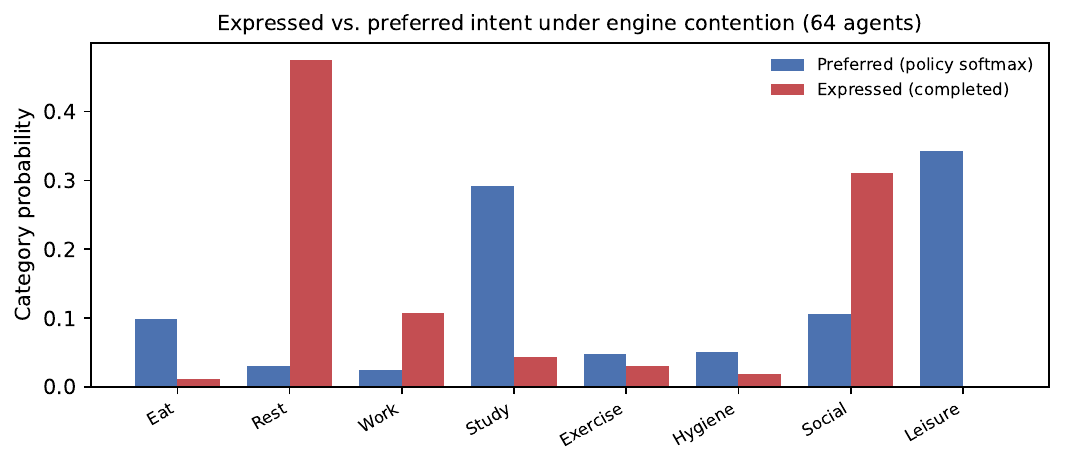}
  \caption{%
    \textbf{Expressed vs.\ preferred intent under engine contention.}
    Policy-preferred category distribution (mean softmax over the
    $20$-d logits, folded to $11$ categories) against the categories
    whose interactions actually complete, aggregated over $64$ agents.
    The policy's top preferences (Leisure, Study) collapse at execution
    while Rest/Social/Work absorb the displaced mass (symmetric KL
    $=9.07$ nats) --- agents are rerouted to reachable affordances.}
  \label{fig:ue5_evp}
\end{figure}

\textbf{Where the contention lands.}
Figure~\ref{fig:ue5_contention} traces zone-capacity utilisation
(occupants$/$capacity) across a representative $64$-agent HybridPCSP
session ($3{,}881$ decisions, $3{,}728$ completed interactions over
$629$\,s). Utilisation is non-uniform: the Rest, Work, Hygiene, and
Exercise zones peak near $0.65$ and Rest sits at a $0.51$ time-mean,
while other zones stay slack. This unevenness is the engine-side
mechanism that compresses the persona signal --- Figure~\ref{fig:ue5_evp}
contrasts the policy's \emph{preferred} intent distribution (mean
softmax over the $20$-d logits, folded to the $11$ affordance
categories) with what is \emph{expressed} (the categories whose
interactions actually complete). The two distributions diverge sharply
(symmetric KL $=9.07$ nats): the policy's largest preferences,
Leisure ($0.34$) and Study ($0.29$), collapse at execution
($0.00$ and $0.04$), while the reachable Rest, Social, and Work
categories absorb the displaced mass ($0.03\!\to\!0.48$,
$0.11\!\to\!0.31$, $0.03\!\to\!0.11$). This is the same ``pushed toward
what is reachable'' effect made concrete: the BT's failure-recovery
decorators reroute an unsatisfiable intent to whichever affordance is
free, attenuating the embedding-to-action pathway that Layer~1 measures
in isolation.

\begin{table}[t]
  \centering
  \small
  \caption{%
    \textbf{BT-abort taxonomy at $64$ agents.} Decision counts and abort
    categories for representative sessions; \texttt{FindBestZone} is the
    contention-induced ``every candidate zone is full'' abort,
    \texttt{zone\_no\_free\_interaction\_point} is the
    selection-to-arrival race, \texttt{path\_follow\_idle\_short} is a
    NavMesh follower stall. The same persona-removal ablation that
    inflates dispersion to $\rho\!=\!0.99$ also moves the failure mass
    from $\sim\!0\%$ into \texttt{FindBestZone}-class aborts, naming
    contention as the engine-side mechanism behind the $\rho$-drop.}
  \label{tab:ue5_failures}
  \setlength{\tabcolsep}{2pt}
  \scriptsize
  \begin{tabularx}{\linewidth}{>{\raggedright\arraybackslash}X r
      >{\raggedright\arraybackslash}p{0.24\linewidth} r r}
    \toprule
    Session / mode & Dec.\ & FindBestZone & ZoneNoFree & PathStall \\
    \midrule
    HybridPCSP (train)        & 2{,}386 & 41   & 0   & 2  \\
    HybridPCSP (held-out, tuned) & 2{,}975 & 0    & 0   & 1  \\
    NoConsist (in-engine)     & 3{,}324 & 0    & 0   & 10 \\
    HybridNoPersona ablation  & ---     & majority of $13.3\%$ aborts & --- & --- \\
    BTOnly (no policy)        & 9{,}644 & 8{,}062 & 85 & 1  \\
    \bottomrule
  \end{tabularx}
\end{table}

\textbf{Engine-side persona--action alignment.}
We additionally test the \PCSPD{} alignment metric inside the engine: for
each session we compute pairwise cosine distance over the $64$-d projected
persona embedding and pairwise symmetric KL over the per-persona mean
softmax(logits), and report Spearman $\rho$ across the $\binom{64}{2}\!=\!2{,}016$
persona pairs. Full PCSP yields engine-side $\rho \in [0.236, 0.257]$ across
two paired sessions, the \texttt{no\_consist} checkpoint yields $\rho = 0.569$,
and \texttt{BTOnly} yields $\rho \approx 0$ (the policy emits zero logits by
construction). All four values are well below the \PCSPD{} research-side
$\rho \approx 0.73$. The hybrid stack itself imposes a ceiling: capacity
contention and the BT's failure-recovery decorators push agents toward what is
reachable, not what their embedding most prefers, compressing the persona
signal at execution time. We treat this gap as a quantified engine--research
mismatch, not a refutation of the in-research result.

Per-decision JSONL traces, the live Gameplay Debugger overlay
(Fig.~\ref{fig:ue5_debugger}), and a $\sim\!2{,}000$-line C++ integration
make the consistency-loss ablation re-runnable inside the engine without
re-instrumenting; full observability instrumentation, trace schema, and
implementation-cost breakdown are described in
App.~\ref{app:ue5_pipeline}.

\subsection{Emergent social structure}
\label{sec:ue5_social}

\begin{figure}[t]
  \centering
  \includegraphics[width=\linewidth]{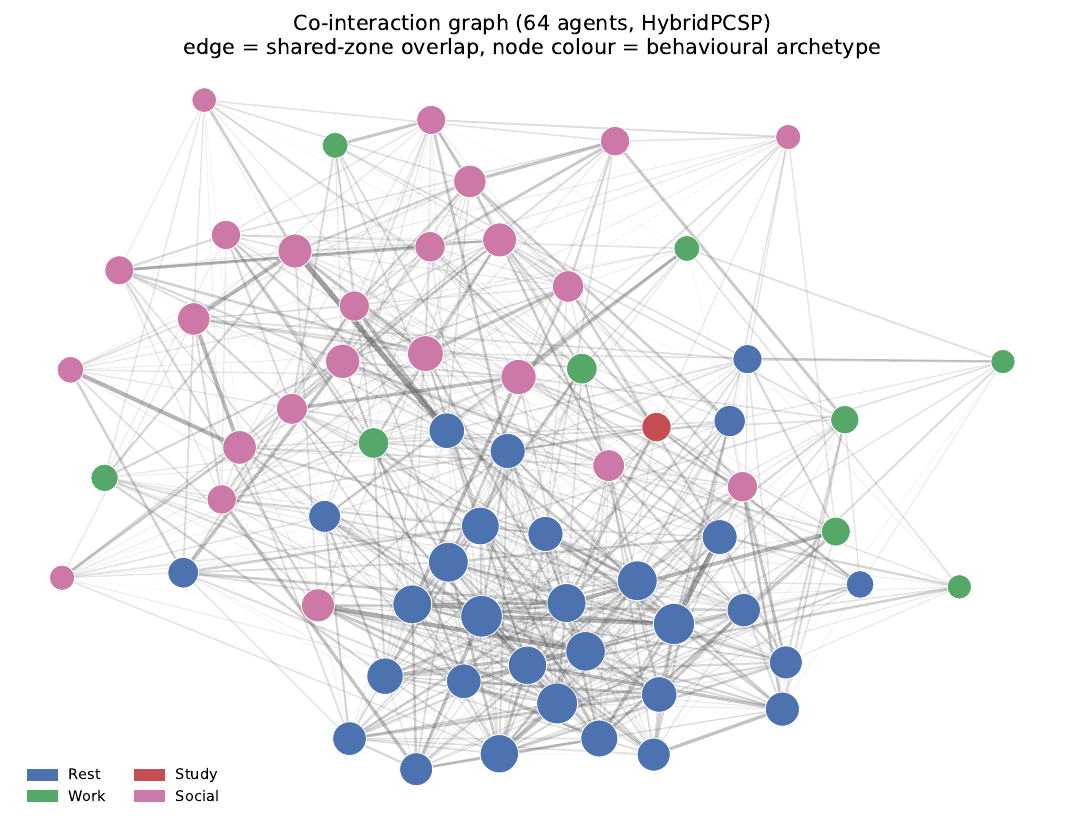}
  \caption{%
    \textbf{Co-interaction graph over the $64$-agent HybridPCSP episode.}
    Nodes are agents, coloured by behavioural archetype (modal expressed
    category); an edge joins two agents whenever their zone-occupancy
    intervals overlap \emph{and} their interaction points lie within
    $250$ world units (same/adjacent seat). Edge opacity and width scale
    with total shared-zone overlap; node size scales with weighted degree.
    Same-archetype agents co-locate well above chance (assortativity
    $0.36$, $63.5\%$ of edges same-archetype).}
  \label{fig:ue5_social}
\end{figure}

The hybrid stack records no explicit social interactions, yet a
co-presence structure emerges purely from $64$ agents independently
choosing where to spend time under a shared world. From the per-agent
traces we reconstruct each agent's zone-occupancy intervals (a
\texttt{decision}$\rightarrow$\texttt{interaction\_complete} pair fixes
a zone category, a time window, and an interaction-point position) and
draw a co-interaction edge between two agents whenever their intervals
overlap in the same zone \emph{and} their interaction points fall within
$250$ world units --- a genuine shared-seat criterion rather than mere
co-membership of a large zone, which a single $36$-capacity zone would
otherwise saturate into a near-complete graph. The resulting graph over
$3{,}728$ completed interactions (Fig.~\ref{fig:ue5_social}) has
$672$ edges (density $0.33$, mean weighted degree $21$) and is
\emph{archetype-assortative}: the attribute assortativity coefficient is
$0.36$ and $63.5\%$ of edges join agents of the same modal archetype,
far above the mixing expected if routine choice were persona-independent.
The effect is robust to the co-location threshold and strengthens
monotonically as the criterion tightens --- assortativity rises
$0.13\rightarrow0.26\rightarrow0.36$ as the radius contracts
$600\rightarrow400\rightarrow250$ units --- confirming that the
structure is driven by genuine physical co-location of like-persona
agents, not by coarse zone coincidence. No social objective, reward
term, or coordination signal is present; the clustering is an emergent
consequence of persona-conditioned activity selection expressed through
the shared engine, and it is the kind of mature deployment-side analysis
that Layers~1 and~2 cannot produce.

\section{Discussion and Future Work}
\label{sec:discussion}

\textbf{Dynamic personas and memory.}
Current \PCSP encodes a static persona text once at NPC creation. Life-simulation
characters should also change: a repeated social conflict may increase
neuroticism, a long friendship may shift agreeableness, and a remembered event
may alter future action preferences. A natural extension is to separate stable
persona traits from dynamic memory state, using a lightweight event-conditioned
update model or retrieval-augmented memory module that can influence the shared
policy without requiring per-step LLM inference~\cite{park2023generative}.
Persona-grounded dialog corpora~\cite{zhang2018personachat} provide a
complementary signal for evolving personas from social interaction.

\textbf{Scaling to richer environments.}
The Melting Pot evidence in \S\ref{sec:meltingpot} now spans three substrates
across distinct social-dilemma categories and includes both held-out-vocabulary
retrieval and cross-substrate transfer (\S\ref{sec:mp_transfer},
Tab.~\ref{tab:mp_transfer}). Two open problems remain. \emph{Held-out
persona recovery} fails consistently---across $11$ full \PCSP{} runs, neither
held-out persona is retrieved at rank~1---reproducing the embedding-margin
condition documented in the companion technical
report~\cite{phase5report}; the natural mechanistic targets are
multi-objective InfoNCE with per-cluster pinning, projection-head re-seeding,
and a controlled-margin held-out corpus. \emph{Cross-substrate transfer}
is positive but asymmetric (CU $\to$ CH $1.79\times$ chance top-1,
CH $\to$ CU at-or-below chance top-1 but $1.39\times$ chance top-3), pointing
to a substrate-invariant projection as the next architectural target.
The UE5 deployment in \S\ref{sec:ue5} shows that the sub-frame inference
profile and the persona-conditioning ablation both survive the move out of
the Python benchmark; what remains is commercial-scale multi-agent
validation in worlds substantially richer than a single district, including
procedurally-generated suites~\cite{cobbe2020procgen} that stress
persona-policy generalization beyond a fixed training map.

\textbf{Human evaluation methodology.}
Automatic metrics are necessary for ablations but insufficient for perceived NPC
believability. Our 30-participant coarse-trace pilot shows above-chance persona
identification, but also many ambiguous or misleading items. Stronger human
evaluation will require an instrument that preserves per-participant responses,
so that variance, confidence, and inter-rater agreement can be analyzed
alongside aggregate accuracy.

\section{Limitations and Scope}
\label{sec:limits}

\PCSP has several limitations that bound the interpretation of the results.

\textbf{Deliberate minimality of Layer 1.}
\PCSPD{} is intentionally simpler than commercial life-simulation worlds.
Behavioral realism is not its purpose; it is the layer at which we can run
controlled InfoNCE ablations across three independent environment
instantiations and thousands of held-out personas. Realism claims in this
paper are grounded in Layers 2 (Melting Pot, \S\ref{sec:meltingpot}) and 3
(UE5, \S\ref{sec:ue5}), where realism, contention, and asynchrony are present
and where the same checkpoint is shown to behave consistently. Commercial-scale
multi-agent validation, continuous physics, longer horizons, asynchronous
events, and richer social affordances beyond Layer~3 remain untested.

\textbf{Action observability.}
The original 12-action space exposes only a thin slice of persona-relevant
behavior. In the 30-participant coarse-trace pilot, aggregate 2AFC accuracy was
above chance, but 13 of 30 items were ambiguous or misleading. Coarse labels
such as \texttt{read}, \texttt{rest}, and \texttt{eat} are difficult to
interpret without temporal, spatial, and social context. \PCSPD-v3 and the
rich rollout renderer address this partly, but finer social and stylistic
action semantics remain needed.

\textbf{Synthetic personas.}
Training personas are generated from Big Five archetypes and occupation
templates. The designer-authored case study in \S\ref{sec:qualcase} tests 50
additional handwritten personas across five sources, but does not establish
robustness to the nuance, inconsistency, or cultural specificity of
production-authored characters.

\textbf{Limited aggregate human evaluation.}
The completed Google Forms pilot collected only item-level A/B selection ratios
for coarse traces. It supports aggregate 2AFC accuracy analysis, but not
participant-level variance, confidence, response time, order effects, or
inter-rater reliability. A more instrumented human study remains future work.

\textbf{Engine--research alignment gap (analysis in
\S\ref{sec:ue5_contention}).}
We do not treat the Layer-1$\to$Layer-3 drop in absolute alignment
($\rho \approx 0.73$ in \PCSPD{} vs.\ $\rho \in [0.236,\,0.257]$ in UE5) as a
method failure. The persona-conditioning ablation and the InfoNCE
consistency-loss replication both remain visible \emph{inside the engine}
(\S\ref{sec:ue5_contention}); the structural cause is engine-side contention,
not loss of method validity. What remains open is whether a less congested
map narrows the absolute gap, and how to apportion the residual between
hybrid-stack ceiling and intrinsic deployment-noise.

\section{Conclusion}
\label{sec:conclusion}

Life simulation games create a scaling challenge that the game AI field has not
yet systematically addressed: how to give hundreds to thousands of NPCs distinct,
consistent, and controllable personalities without proportional authoring or
inference cost. We show that \emph{persona-conditioned shared
policies}---a single RL policy conditioned on frozen LLM persona
embeddings---can satisfy the four axes that matter for practical deployment:
persona consistency, natural-language controllability, zero-shot
generalization, and real-time inference.

Across three \PCSPD{} settings, \PCSP achieves up to 17$\times$ above-chance
\emph{compositional} zero-shot persona identification (unseen-occupation
held-out crosses), $\rho\!\approx\!0.73$ semantic-behavioral alignment, and
22$\times$ faster inference than an LLM-as-policy baseline. We separate this
regime from \emph{vocabulary-expansion} held-out evaluation in Melting Pot
(\S\ref{sec:mp_transfer}), where top-1 retrieval on the two genuinely held-out
persona tokens remains at $0$ across $11$ runs and which we flag as the most
direct open problem.
External validation on three \textit{Melting Pot 2.4.0} substrates
(\texttt{commons\_harvest\_\_open}, \texttt{clean\_up},
\texttt{prisoners\_dilemma\_in\_the\_matrix\_\_repeated}) confirms that the same
method produces persona-conditioned behavioral divergence across commons-pool,
public-good, and dyadic-matrix social dilemmas, with the no-InfoNCE ablation
collapsing trajectory$\to$persona retrieval to chance in every substrate. The
main empirical finding is that the InfoNCE trajectory-consistency objective is
load-bearing: removing it collapses zero-shot persona traceability to chance
even when reward improves. The next steps are to couple static personas with
dynamic memory, test cross-substrate persona transfer and held-out-persona
recovery against the mechanistic targets identified in the companion technical
report~\cite{phase5report}, and complete human studies using rich trajectory
traces.

\renewcommand{\topfraction}{0.95}
\renewcommand{\bottomfraction}{0.95}
\renewcommand{\textfraction}{0.05}
\renewcommand{\floatpagefraction}{0.80}
\renewcommand{\dbltopfraction}{0.95}
\renewcommand{\dblfloatpagefraction}{0.80}
\setcounter{topnumber}{4}
\setcounter{bottomnumber}{4}
\setcounter{totalnumber}{8}
\appendices
\section{Layer 1 base-scale (v1), expanded-scale (v2), and v3-large replications}
\label{app:v1v2}

The v3 results in \S\ref{sec:evidence} are the primary Layer-1 evidence. We
include here the v1 (base-scale, 12-action), v2 (expanded-scale, 12-action),
and v3-large (expanded-scale, 20-action) replications that establish the
InfoNCE consistency objective as load-bearing across grid size, agent count,
persona-set size, and action ontology. The protocols are described in the
Experimental Setup paragraphs of \S\ref{sec:evidence}; metrics, baselines,
and chance levels are unchanged. v3-large is the only one of the three with
3-seed aggregation; v1 and v2 report point estimates from a single seed.

\begin{table}[!t]
  \centering
  \small
  \caption{%
    \textbf{v1 results} (6$\times$6, 4 agents, 300 personas).
    ZS Acc: zero-shot $k$-NN accuracy on 60 unseen personas
    (random $\approx\!1.7\%$). $\rho$: Spearman (projected persona dist.\ vs.\ KL).}
  \label{tab:results_v1}
  \setlength{\tabcolsep}{3pt}
  \begin{tabular}{lrrrr}
    \toprule
    Model & Reward $\uparrow$ & ZS Acc $\uparrow$ & $\rho$ $\uparrow$ & Lat.\ $\downarrow$ \\
    \midrule
    \textbf{\PCSP (full)}  & 83.4 & \textbf{0.193} & \textbf{0.728} & 1.97\,ms \\
    PCSP (no\_consist)     & 84.3 & 0.017*  & 0.638 & 2.03\,ms \\
    PCSP (no\_diverse)     & 76.8 & 0.147   & ---$^\dagger$ & 1.79\,ms \\
    B1 No-Persona          & 73.2 & ---     & ---   & 1.83\,ms \\
    B3 SBERT               & \textbf{86.2} & --- & --- & 1.88\,ms \\
    B5 LLM-policy          & ---  & ---     & ---   & 43.7\,ms \\
    \bottomrule
  \end{tabular}\\[2pt]
  {\footnotesize * Near-random (failure mode).
    $^\dagger$KL $\approx\!0.39$; $\rho$ not meaningful.}
\end{table}

\begin{table}[!t]
  \centering
  \small
  \caption{%
    \textbf{v2 results} (12$\times$12, 16 agents, 500 personas).
    ZS Acc on 100 unseen personas (random $=\!1.0\%$).
    Lower absolute reward reflects the harder, larger environment.}
  \label{tab:results_v2}
  \setlength{\tabcolsep}{3pt}
  \begin{tabular}{lrrrr}
    \toprule
    Model & Reward $\uparrow$ & ZS Acc $\uparrow$ & $\rho$ $\uparrow$ & KL $\uparrow$ \\
    \midrule
    \textbf{\PCSP (full)}  & 64.2 & 0.023 & \textbf{0.725} & \textbf{5.40} \\
    PCSP (no\_consist)     & 50.1 & 0.000*  & 0.253 & 16.47$^\ddagger$ \\
    PCSP (no\_diverse)     & 71.6 & 0.013   & ---$^\dagger$ & 0.48 \\
    PCSP (concat)          & \textbf{74.8} & \textbf{0.027} & ---$^\dagger$ & 1.34 \\
    \bottomrule
  \end{tabular}\\[2pt]
  {\footnotesize * Exactly 0 (total failure). $^\dagger$KL too low; $\rho$ not meaningful.
    $^\ddagger$High KL without consistency reflects unstructured divergence, not persona alignment.}
\end{table}

\begin{table}[!t]
  \centering
  \small
  \caption{%
    \textbf{v3-large results} (12$\times$12, 16 agents, 500 personas,
    20-action ontology; 400 train / 100 zero-shot test). 3 seeds per arm
    (mean $\pm$ std).
    ZS Acc on 100 unseen personas (random $=\!1.0\%$).
    Coh.: trajectory coherence ratio (intra/inter persona cosine similarity).}
  \label{tab:results_v3_large}
  \setlength{\tabcolsep}{3pt}
  \begin{tabular}{lrrr}
    \toprule
    Model & Reward $\uparrow$ & ZS Acc $\uparrow$ & Coh.\ $\uparrow$ \\
    \midrule
    \PCSP (full)         & $119.0\pm2.0$  & $0.040\pm0.009$  & $1.89\pm0.18$ \\
    \PCSP (no\_consist)  & $122.2\pm0.9$  & $0.013\pm0.005$*  & $1.05\pm0.01$ \\
    \PCSP (no\_diverse)  & $112.8\pm8.5$  & $0.054\pm0.029$  & $2.19\pm0.39$ \\
    \textbf{\PCSP (concat)} & $120.1\pm0.9$ & \textbf{$0.058\pm0.004$} & \textbf{$2.36\pm0.47$} \\
    B1 No-Persona        & $57.8\pm1.1$   & ---              & --- \\
    B3 SBERT             & $119.4\pm1.3$  & ---              & --- \\
    \bottomrule
  \end{tabular}\\[2pt]
  {\footnotesize * At random chance (0.010). The consistency loss is held at
   zero during training, so trajectory and persona embeddings receive no
   alignment signal; coherence ratio collapses to 1.05 (intra $\approx$
   inter), reproducing the v3-base finding at 4$\times$ grid, 4$\times$
   agents, 5/3$\times$ personas, under the 20-action ontology.}
\end{table}

\section{UE5 System Architecture}
\label{app:ue5_system}

\begin{figure}[t]
  \centering
  \includegraphics[width=\linewidth]{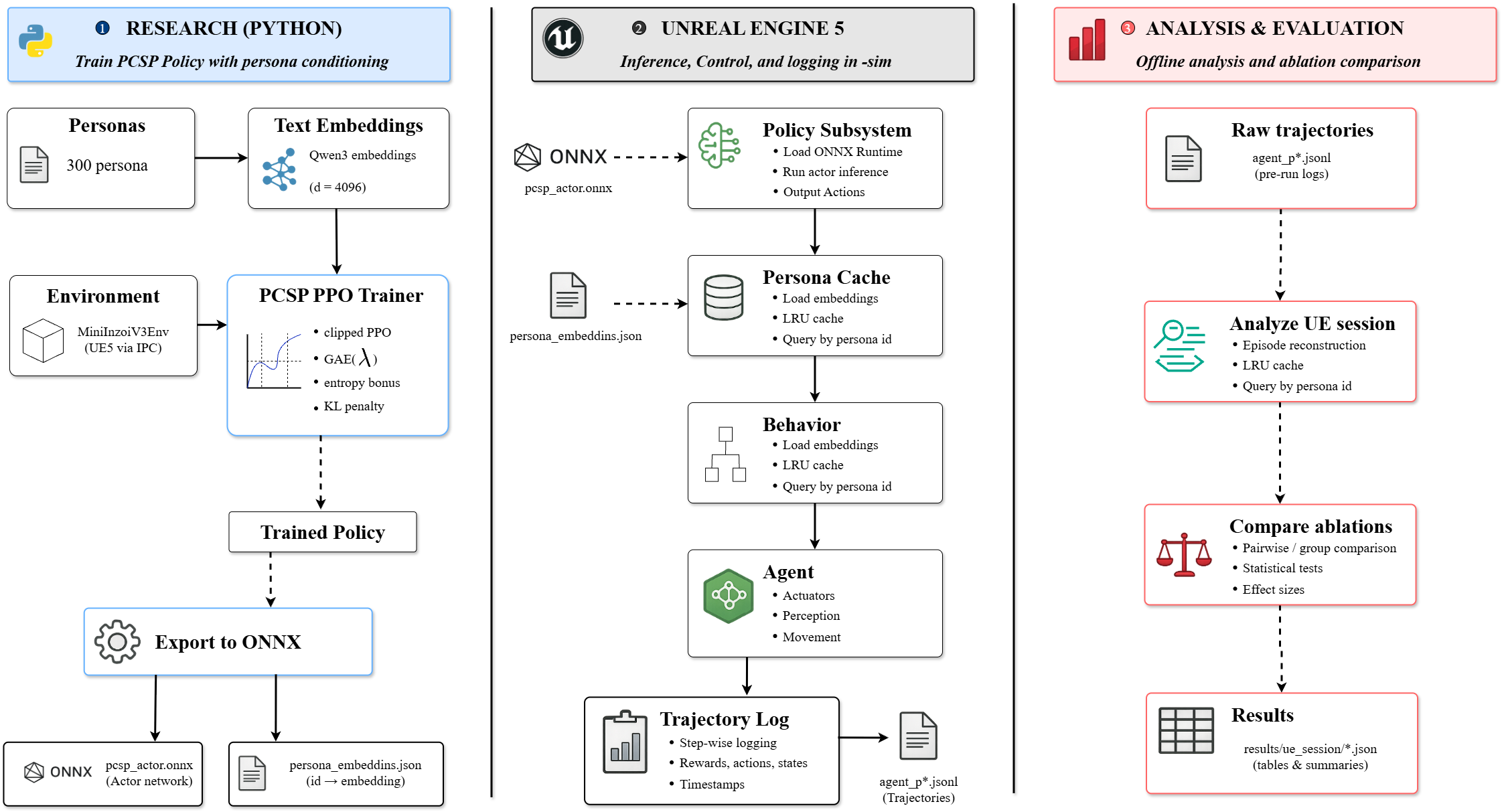}
  \caption{%
    \textbf{Research--UE5 split.} The Python pipeline (left) trains
    \PCSP and exports a single ONNX actor plus a JSON of L2-normalised
    persona projections. Unreal (centre) loads both into
    \texttt{UPCSPPolicySubsystem} and runs synchronous inference inside a
    Blackboard-driven Behaviour Tree; per-decision JSONL traces feed the
    offline analysis scripts (right). No engine-side training.
  }
  \label{fig:ue5_system}
\end{figure}

\begin{figure}[t]
  \centering
  \includegraphics[width=\linewidth]{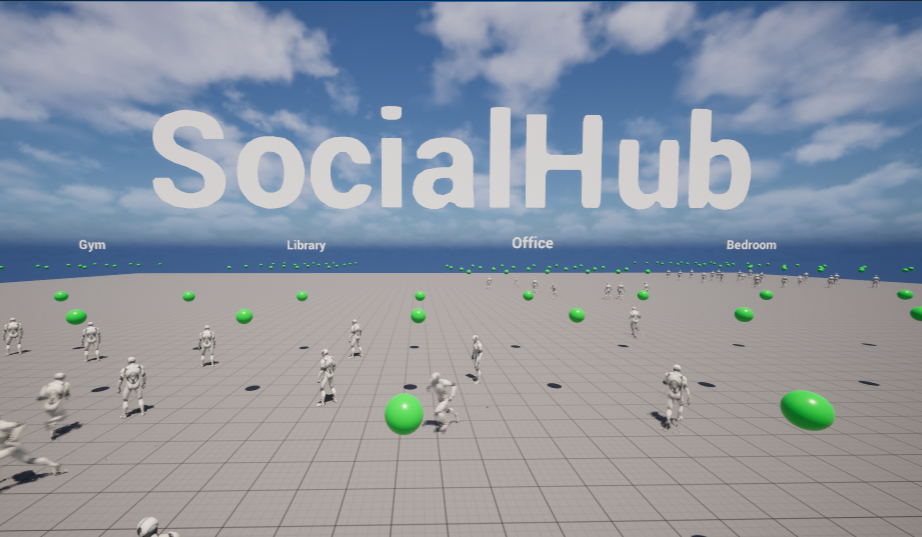}
  \caption{%
    \textbf{\texttt{Map\_PCSPDistrict\_M} in PIE.} Top-down view with
    affordance-zone labels (SocialHub, Library/Study, Office/Work,
    Bedroom/Rest, \ldots); each green sphere is an
    \texttt{APCSPAgentCharacter} executing the hybrid PCSP/BT stack. The
    map has $10$ zones spanning the v3 affordance taxonomy (Kitchen,
    Bedroom, Office, Library, Gym, Bathroom, SocialHub, Park, Shop;
    Leisure folded into Observe).
  }
  \label{fig:ue5_screenshot}
\end{figure}

\begin{figure}[t]
  \centering
  \includegraphics[width=\linewidth]{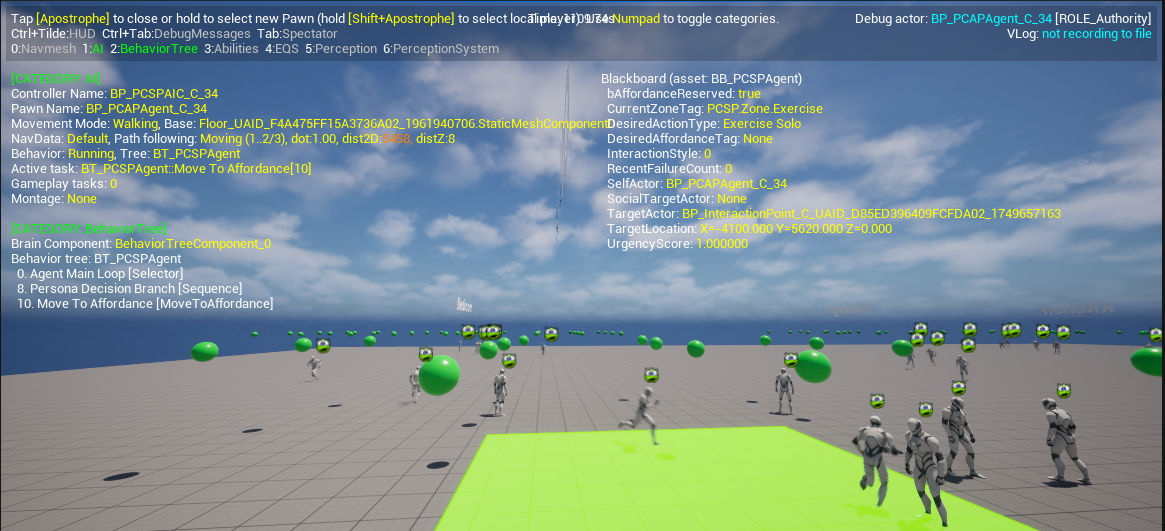}
  \caption{%
    \textbf{Gameplay Debugger overlay on a single agent.}
    Yellow text (right) is the live Blackboard / BT snapshot for the
    selected \texttt{APCSPAgentCharacter}: current
    \texttt{DesiredActionType}, \texttt{UrgencyScore},
    \texttt{bAffordanceReserved}, and the active BT node. This view
    confirms that the persona branch is routing decisions through
    \texttt{BTTask\_PCSPDecision} $\to$ \texttt{MoveToAffordance} $\to$
    \texttt{PerformInteraction} and that the emergency decorator fires
    only when \texttt{UrgencyScore}~$\geq\!0.85$.
  }
  \label{fig:ue5_debugger}
\end{figure}

This appendix collects the engine-integration scaffolding for the
Layer-3 deployment whose \emph{scientific} findings are reported in
\S\ref{sec:ue5}. Figure~\ref{fig:ue5_system} shows the
research\,/\,engine split: the Python pipeline trains \PCSP and exports a
single ONNX actor plus a JSON of L2-normalised persona projections;
Unreal loads both into \texttt{UPCSPPolicySubsystem} and runs synchronous
inference inside a Blackboard-driven Behaviour Tree; per-decision JSONL
traces feed the offline analysis scripts. No engine-side training.
Figure~\ref{fig:ue5_screenshot} shows \texttt{Map\_PCSPDistrict\_M} in PIE
with the affordance-zone layout used throughout the Layer-3 evaluation,
and Fig.~\ref{fig:ue5_debugger} shows the live Gameplay Debugger overlay
that exposes per-agent Blackboard and BT state at runtime.

\section{Inference Pipeline, Observability, and Implementation Cost}
\label{app:ue5_pipeline}

This appendix documents the engine-side instrumentation and the
integration footprint that supports the Layer-3 findings in
\S\ref{sec:ue5}. The key claim is reproducibility: the same
consistency-loss ablation that defines the paper's central Layer-1
finding can be re-run end-to-end inside the engine without
re-instrumenting, on top of an integration that is small enough to be
audited in a single sitting.

\textbf{Observability.}
In addition to the offline JSONL traces, the live Gameplay Debugger
(Fig.~\ref{fig:ue5_debugger}) exposes per-agent Blackboard and BT state
at runtime. Every agent emits a per-decision JSONL trace (action,
$20$-d softmax logits, $8$-d needs snapshot, urgency, reward, zone
selection diagnostic, failure reason). The same
\texttt{analyze\_ue\_session.py} pipeline that aggregates these traces
also computes per-persona symmetric KL between paired PIE sessions, so
the consistency-loss ablation that defines the paper's central finding
can be re-run inside the engine without re-instrumenting.

\textbf{Implementation cost.}
The engine integration is deliberately lightweight: $\sim\!2{,}000$
lines of C++ across \texttt{Source/cnzoi/PCSP/} (one world subsystem,
five agent components, three BT tasks, an agent character), one
$\sim\!4$\,MB ONNX file plus a $\sim\!150$\,KB persona-embedding JSON,
and zero engine-side training. The Python pipeline remains the single
source of truth for both the policy and the persona embeddings. The
64-agent stress test runs at $\geq\!60$\,FPS in PIE on a single consumer
GPU/CPU; inference is synchronous on the BT tick with a $0.5$\,s
decision throttle and exponential backoff on repeated move failures,
capping ONNX calls at $\leq\!128/\text{s}$ at this scale---comfortably
within the NNERuntimeORT CPU path's headroom without async batching.

\section{Realtime Scaling Sweep}
\label{app:ue5_scaling}

This appendix reports the single-workstation scaling sweep that
establishes the realtime envelope of the hybrid stack and identifies
NavMesh \texttt{FindPath} saturation (not policy inference) as the hard
ceiling. Table~\ref{tab:ue5_scaling} is the source data behind the
``\,$n\!\leq\!64$ recommended, $n\!=\!96$ soft cap, $n\!\geq\!128$
requires async batched pathfinding\,'' guidance referenced in
\S\ref{sec:ue5}.

\textbf{Protocol.}
$18$ runs total
(\texttt{research/\allowbreak results/\allowbreak ue\_sessions/\allowbreak scaling\_20260520/}).
All runs share the same ONNX policy, the same $0.5$\,s decision throttle
with urgency-driven preemption, and a $60$\,FPS wall budget.
The sweep is fully automated by \texttt{tools/run\_scaling\_sweep.ps1},
which sets \texttt{-PCSP\_AgentCount}\,/\,\allowbreak\texttt{SpawnSeed}\,/\,\allowbreak\texttt{RunDurationSeconds}
on the command line (\texttt{FParse::Value} reads these before any
\texttt{BeginPlay}); auto-quit fires from \texttt{FTSTicker::GetCoreTicker}
inside \texttt{UPCSPPerfSamplerSubsystem}, and a $+90$\,s PowerShell
watchdog guarantees forward progress.

\textbf{Observations.}
(i) \emph{ONNX inference is not the bottleneck.} Mean per-call latency
stays at $183$--$202$\,\textmu s through $n{=}64$; the apparent drop to
$154$\,\textmu s ($n{=}96$) and $132$\,\textmu s ($n{=}128$) reflects
CPU-scheduler timeslicing at saturation, not model speedup.
(ii) \emph{Frame time scales near-linearly at $\sim\!0.27$\,ms/agent},
with the $p95$ frame budget intact through $n{=}96$ ($15.9$\,ms) and only
$0.4$\,ms over the $16.67$\,ms $60$\,FPS budget at $n{=}128$.
(iii) \emph{NavMesh pathfinding is the hard ceiling.} BT-abort failure
rate is $\leq\!0.2\%$ for $n{\leq}64$, rises to $4.7\%$ at $n{=}96$, and
collapses to $\mathbf{44.9\%}$ at $n{=}128$ as concurrent
\texttt{FindPath} requests exceed the Recast query queue's async capacity
(synchronous BT \texttt{MoveToAffordance}). The recommended real-time
operating point is $\boldsymbol{n{\leq}64}$; $96$ is a soft cap;
$128{+}$ requires async batched pathfinding or a crowd-simulation
fallback. Intent throughput is stable at $5.6$--$6.1$ intents/agent/min
for $n{\leq}64$ --- per-agent decision rate is invariant under crowd
scaling.

\begin{table}[t]
  \centering
  \small
  \caption{%
    \textbf{UE5 scaling sweep (Fig.~\ref{fig:ue5_screenshot}; $18$ runs).}
    $\{8,16,32,64,96,128\}$ agents $\times\;3$ seeds $\times\;630$\,s on
    \texttt{Map\_PCSPDistrict\_M} in standalone \texttt{-game} mode.
    All runs use the same frozen v3 ONNX policy, identical zone
    capacities, $0.5$\,s decision throttle, and $60$\,FPS wall budget.
    Inference latency stays well under the $250$\,\textmu s per-agent budget
    through $n{=}64$; the hard ceiling is NavMesh \texttt{FindPath}
    saturation at $n{\geq}96$, not policy inference.
    Source: \texttt{latency\_budget.tsv} from
    \texttt{research/scripts/analyze\_scaling\_sweep.py}.}
  \label{tab:ue5_scaling}
  \setlength{\tabcolsep}{2pt}
  \scriptsize
  \begin{tabular}{rrrrrrr}
    \toprule
    Agents & infer\,$\bar{\mu}$s & infer\,$p95\,\mu$s & frame\,$\bar{\text{ms}}$ & frame\,$p95$\,ms & Fail\,\% & Int./ag./min \\
    \midrule
    8   & 183.2 & 234.8 &  5.57 &  7.75 &  0.1 & 5.67 \\
    16  & 184.1 & 230.5 &  5.84 &  8.19 &  0.0 & 5.67 \\
    32  & 202.6 & 257.6 &  7.53 & 11.27 &  0.0 & 6.06 \\
    \textbf{64}  & \textbf{199.8} & \textbf{264.1} & \textbf{10.25} & \textbf{13.38} & \textbf{0.2} & \textbf{5.61} \\
    96  & 153.7 & 198.1 & 11.62 & 15.89 &  4.7 & 5.35 \\
    128 & 132.0 & 181.6 & 14.39 & 17.05 & \textbf{44.9} & 4.98 \\
    \bottomrule
  \end{tabular}
\end{table}

\clearpage
\bibliographystyle{IEEEtran}
\bibliography{refs}

\end{document}